\documentclass{bmvc2k}

\usepackage{multirow}
\usepackage{multicol}
\usepackage{booktabs}
\usepackage{graphicx}
\usepackage{xcolor}
\usepackage{wrapfig}
\usepackage{amssymb}
\usepackage{amsmath}
\usepackage{pifont}
\usepackage{capt-of}
\usepackage{colortbl}
\usepackage{hyperref}

\hypersetup{
    colorlinks=true,
    urlcolor=magenta
}

\newcommand{\cmark}{\ding{51}}%
\newcommand{\xmark}{\ding{55}}%

\title{With a Little Help from my Temporal Context: Multimodal Egocentric Action Recognition}

\addauthor{Evangelos Kazakos}{Evangelos.Kazakos@bristol.ac.uk}{1}
\addauthor{Jaesung Huh}{jaesung@robots.ox.ac.uk}{2}
\addauthor{Arsha Nagrani$\dagger$}{arsha.nagrani@gmail.com}{2}
\addauthor{Andrew Zisserman}{az@robots.ox.ac.uk}{2}
\addauthor{Dima Damen}{Dima.Damen@bristol.ac.uk}{1}

\addinstitution{
 Dept. of Computer Science\\
 University of Bristol\\
 Bristol, UK
}
\addinstitution{
 Visual Geometry Group\\
 University of Oxford,\\
 Oxford, UK
}

\runninghead{Kazakos et al.}{With a Little Help from my Temporal Context}

\begin{document}

\maketitle

\begin{abstract}
In egocentric videos, actions occur in quick succession.
We capitalise on the action's temporal context and propose a method that learns to attend to surrounding actions in order to improve recognition performance.
To incorporate the temporal context, we propose a transformer-based multimodal model that ingests video and audio as input modalities, with an explicit language model providing action sequence context to enhance the predictions.
We test our approach on EPIC-KITCHENS and EGTEA datasets reporting state-of-the-art performance. Our ablations showcase the advantage of utilising temporal context as well as incorporating audio input modality and language model to rescore predictions. Code and models at: \url{https://github.com/ekazakos/MTCN}.
\end{abstract}


\renewcommand*{\thefootnote}{\fnsymbol{footnote}}
\footnotetext[2]{Now at Google Research.}
\renewcommand*{\thefootnote}{\arabic{footnote}}

\section{Introduction}
\label{sec:intro}

\begin{figure}[t]
    \centering
    \includegraphics[width=\textwidth]{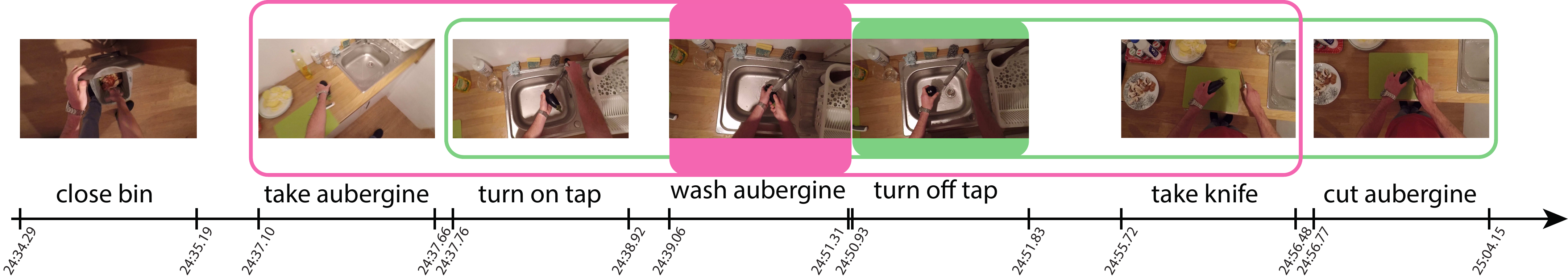}
   \vspace*{-16pt}
    \caption{Egocentric video demonstrating two temporal context windows (pink, green), centred around the action to be recognised. We can infer `wash aubergine' with higher accuracy if we know that the tap was turned on before and turned off afterwards.} 
    \vspace*{-12pt}
    \label{fig:temporal_context_example}
\end{figure}

Action recognition in egocentric video streams from sources like EPIC-KITCHENS poses a number of challenges that differ substantially from those of conventional third-person action recognition -- where training and evaluation is on 10 second video clips and classes are quite high-level~\cite{kay2017kinetics}.
Actions are fine-grained (e.g.\ `open bottle') and noticeably short, often one second or shorter.
Along with the challenge, the footage offers an under-explored opportunity, as actions are captured in long untrimmed videos of well-defined and at-times predictable sequences. For example the action `wash aubergine' can be part of the following sequence -- you first `take the aubergine', `turn on the tap', `wash the aubergine' and finally `turn off the tap' (Fig.~\ref{fig:temporal_context_example}). Furthermore, the objects (the aubergine and tap in this case) are persistent over some of the neighbouring actions.

In this work, we investigate utilising not only the action's temporal context in the data stream, but also the temporal context from the labels of neighbouring actions.
We propose a model that attends to neighbouring actions\footnote{Note that these action start/end times are readily available in labelled datasets for untrimmed videos and do not require additional labels. We just leverage these.}. 
Concretely, we use
the attention mechanism of a multimodal transformer architecture to take account of the context in both the data and labels, using three modalities: vision, audio and language. 
We are motivated by previous works that demonstrate the significance of audio in recognising egocentric actions~\cite{kazakos2019TBN,Wang_2020_CVPR,xiao_2020_arxiv}, and thus include the auditory temporal context in addition to the visual clips. We also utilise context further, by training a language model on the sequence of action labels, inspired by the success of language models \cite{devlin2018bert,yang2019xlnet, brown2020language} in re-scoring model outputs for speech recognition~\cite{chan2016listen, mikolov2010recurrent, hannun2014deep} and machine translation~\cite{gulcehre2017integrating} (Fig. \ref{fig:av_transformer}).

Our main contributions are summarised as follows: First, we formulate temporal context as a sequence of actions surrounding the action in a sliding window. Second, we propose a novel framework able to model multimodal temporal context. It consists of a transformer encoder that uses vision and audio as input context, and a language model as output context operating on the action labels. 
Third, we obtain state-of-the-art performance on two datasets: (i) the large-scale egocentric dataset EPIC-KITCHENS, outperforming high-capacity end-to-end transformer models; and (ii) the EGTEA dataset. Finally, we include an ablation study analysing the importance of the
extent of the temporal context and of the various modalities.

\vspace*{-10pt}
\section{Related Work}
\label{sec:related}
\vspace*{-2pt}

\noindent\textbf{Action Recognition}. 
There is a rich literature of seminal works in action recognition 
innovating temporal sampling~\cite{TSN2016ECCV,zhou2018temporal,kazakos2019TBN}, multiple streams~\cite{Feichtenhofer_2019_ICCV}, spatio-temporal modelling~\cite{Tran_2018_CVPR,lin2019tsm,Girdhar_2019_CVPR} or modelling actions as transformations from initial to final states~\cite{Wang_Transformation}.
Our work is related to the more recent transformer-based approaches~\cite{Girdhar_2019_CVPR,Neimark_2021_arXiv,Bertasius_2021_arXiv,Arnab_2021_arXiv,bulat_2021_arXiv,patrick_2021_arXiv,Nagrani21c}.
We compare our model with four recent works~\cite{Arnab_2021_arXiv,bulat_2021_arXiv,patrick_2021_arXiv,Nagrani21c}. 
\cite{Arnab_2021_arXiv} investigated spatio-temporal attention factorisation schemes, while in \cite{bulat_2021_arXiv} the authors propose full-attention within a temporal window. \cite{patrick_2021_arXiv} proposes temporal attention along trajectories with learnt tokens. \cite{Nagrani21c} proposes fusion bottlenecks for cross-modal attention. These approaches require training on large-scale datasets and strong data augmentation for generalisation. Our model operates on pre-extracted features which are processed with a lightweight transformer and outperforms these works by relying on multimodal temporal context. 

In egocentric action recognition, researchers propose a range of techniques to address its unique challenges \cite{Sudhakaran_2019_CVPR,Baradel_2018_ECCV,Li_2021_CVPR,kazakos2019TBN,Munro_2020_CVPR}. \cite{Sudhakaran_2019_CVPR} considered long-term understanding through an LSTM with attention to focus on relevant spatio-temporal regions, but the approach operates within the action clip solely.
\cite{Baradel_2018_ECCV} shows that modelling hand-object interactions is beneficial, while in \cite{Li_2021_CVPR} the authors pre-train egocentric networks by distilling egocentric signals from large-scale third-person datasets. \cite{kazakos2019TBN} shows that audio is key to egocentric action recognition due to sounds produced from interactions with objects. In \cite{Munro_2020_CVPR},
the authors propose a multimodal unsupervised domain adaptation approach to tackle the distribution shift between environments. We build on these works but propose the first approach to incorporate both audio-visual and language-model predictions.

\noindent\textbf{Temporal Context for Video Recognition Tasks}.
A few works have considered temporal context to improve the models' performance for action anticipation~\cite{Sener_2020_ECCV,Furnari_2019_ICCV}, action recognition \cite{Wu_2019_CVPR,zhang_2021_CVPR,Ng_2019_arXiv,Cartas_2021_ICPR}, and object detection \cite{Beery_2020_CVPR,Bertasius_2020_CVPR}.
\cite{Sener_2020_ECCV} proposes a model that operates on multi-scale past temporal context for action anticipation while they also test on action recognition by modifying the architecture to consider both past and future context.
In \cite{zhang_2021_CVPR}, a set of learnable query vectors attends to dense temporal context to identify events in untrimmed videos. In \cite{Ng_2019_arXiv}, an encoder-decoder LSTM is proposed for classifying sequences of human actions, effectively attending to the relevant temporal context of each action. 
Closest to our work is \cite{Wu_2019_CVPR}, where long-term features from the past and the future of an untrimmed video are utilised, to improve the recognition performance of the ongoing action. A key difference is that we exploit both the temporal bounds and predicted labels of neighbouring actions.

\noindent\textbf{Multimodal transformers}.
The self attention mechanism of transformers provides a natural bridge to connect multimodal signals. 
Applications include audio enhancement~\cite{ephrat2018looking,tzinis2020into}, speech recognition~\cite{harwath2017unsupervised}, image segmentation~\cite{ye2019cross,tzinis2020into}, cross-modal sequence generation~\cite{gan2020foley,li2021learn,li2020learning}, video retrieval~\cite{gabeur2020multi} and image/video captioning/classification~\cite{lu2019vilbert,sun2019videobert, sun2019learning, li2019entangled, iashin2020multi,jaegle2021perceiver}. A common paradigm (which we also adapt) is to use the output representations of single modality convolutional networks as inputs to the transformer~\cite{lee2020parameter,gabeur2020multi}. Unlike these works, we use transformers to combine modalities in two specific ways -- we first combine audio and visual inputs to predict actions based on neighbouring context, and then \textit{re-score} these predictions with the help of a language model applied on the outputs. 

\noindent\textbf{Language models for action detection}.
There are a few works that incorporate a language model in action detection. The most relevant work is \cite{richard2016temporal}, which utilises a statistical n-gram language model along with a length model and an action classifier. \cite{lin2017ctc} combines the Connectionist Temporal Classification (CTC)~\cite{graves2006connectionist} with a language model to learn relationships between actions. Both of these works improve the results by considering the contextual structure of the sequence of actions, albeit relying on a statistical language model, whereas in this work we utilise a neural language model which has shown better performance~\cite{bengio2003neural, doval2019comparing, kim2016character}.

\vspace*{-10pt}
\section{Multimodal Temporal Context Network (MTCN)}
\label{sec:method}
\vspace*{-2pt}

Given a long video, we predict the action in a video segment by leveraging the \emph{temporal context} around it. 
We define the temporal context as the sequence of neighbouring actions that precede and succeed the action, 
and aim to leverage that information, when useful, through learnt attention.
We utilise multimodal temporal context both at the input and the output of our model. 
An audio-visual transformer ingests a temporally-ordered sequence of visual inputs, along with the corresponding sequence of auditory inputs.
We use modality-independent positional encodings as well as modality-specific encodings. 
The language model, acting on the output of the transformer learns the prior temporal context of actions, i.e.\ the probability of the sequence of actions, using a learnt text embedding space.

Inspired by similar approaches~\cite{Wray_2019_ICCV}, and instead of using a single summary embedding as in prior works for image \cite{dosovitskiy2020image} and action classification \cite{Arnab_2021_arXiv}, the audio-visual model utilises two separate summary embeddings to attend to the action (i.e.\ verb) class and the object (i.e.\ noun) class.
This allows the model to attend independently to the temporal context of verbs vs objects. 
For example, the object is likely to be the same in neighbouring actions while the possible sequences of verbs can be independent of objects (e.g.\ `take' $\rightarrow$ `put').
Each summary embedding uses a different learnt classification token, and the classifier predicts a verb and a noun from the summary embeddings.
The predictions of the audio-visual transformer are then enhanced by filtering out improbable sequences using the language model. We term the proposed model Multimodal Temporal Context Network (MTCN).

\begin{figure}[t]
    \centering
    \includegraphics[width=\textwidth]{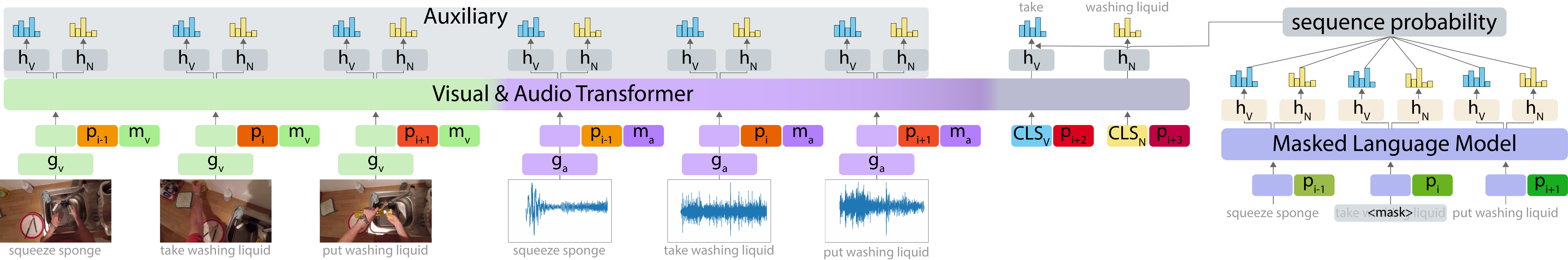}
    \vspace*{-16pt}
    \caption{The Multimodal Temporal Context Network (MTCN). Visual and auditory tokens are tagged with  positional and modality encodings.  An audio-visual transformer encoder attends to the sequence. Verb and noun summary embeddings with independent positional encodings, predict the action at the centre of the window (`take washing liquid'). The classifier also predicts the sequence of actions to train an auxiliary loss that enhances the prediction of the centre action. The language model filters out improbable sequences.} 
    \vspace*{-16pt}
    \label{fig:av_transformer}
\end{figure}

In the next three subsections, we detail the architectural components of MTCN, as well as our training strategy. 
An overview of MTCN is depicted in Fig.~\ref{fig:av_transformer}.

\vspace*{-10pt}
\subsection{Audio-Visual Transformer}
\label{subsec:sv_transformer}
\vspace*{-2pt}

Let $X_v\in\mathbb{R}^{{w} \times d_v}$ be the sequence of visual inputs from a video, and $X_a\in\mathbb{R}^{w\times d_a}$ the corresponding audio inputs\footnote{Video and audio inputs/features are synchronised, therefore they have the same length $w$.}, for $w$ consecutive actions in the video (i.e.\ the temporal context window), with $d_v\text{, }d_a$ being the input dimensions of the two modalities respectively.
$X_v$ and $X_a$ correspond to features extracted from visual and auditory networks, respectively.
Our temporal window is centred around an action $b_i$ with surrounding action segments, excluding any background frames. That is, each action $b_j$ within the window, $i-\frac{w-1}{2}\leq j \leq i+\frac{w-1}{2}$ is part of the transformer's input.

\noindent\textbf{Encoding layer}. 
Our model first projects the inputs $X_{v}\text{, }X_{a}$ to a lower dimension $D$  and tags each with positional and modality encodings.
Then, an audio-visual encoder performs self-attention on the sequence to aggregate relevant audio-visual temporal context from neighbouring actions. 
Because all self-attention operations in a transformer are permutation invariant,  we use positional encodings to retain information about the ordering of actions in the sequence.
We use $w$ learnt absolute positional encodings, shared between audio-visual features to model corresponding inputs from the two modalities. 
Modality encodings, $m_v\text{, }m_a\in\mathbb{R}^D$, are learnt vectors added to discriminate between audio and visual tokens. 

A classifier predicts the action $b_i$, using two summary embeddings, acting on the learnt verb/noun tokens. 
We use the standard approach of appending learnable classification tokens to the end of the sequence but use two tokens, one for verbs and one for nouns, denoted as $\text{CLS}_{\text{V}}\text{, }\text{CLS}_{\text{N}}\in\mathbb{R}^D$, with unique positional encodings.
To summarise, the encoding layer transforms the inputs $X_v$ and $X_a$ as follows:
\vspace*{-6pt}
\begin{gather}
    X_{v_j}^e = g_v(X_{v_j})+p_j+m_v  \qquad   X_{a_j}^e = g_a(X_{a_j})+p_j+m_a \qquad \forall j\in[1,...,w]\\
    \text{CLS}_{\text{V}}^e = \text{CLS}_{\text{V}} + p_{w+1}   \qquad   \text{CLS}_{\text{N}}^e = \text{CLS}_{\text{N}} + p_{w+2} \qquad  
    X^e=[X_v^e;X_a^e;\text{CLS}_{\text{V}}^e;\text{CLS}_{\text{N}}^e]\text{, }
\end{gather}
where $[;]$ denotes input concatenation and $p\in\mathbb{R}^{(w+2)\times D}$ are the positional encodings. $g_v(\cdot):\mathbb{R}^{d_v}\mapsto\mathbb{R}^{D}$ and $g_a(\cdot):\mathbb{R}^{d_a}\mapsto\mathbb{R}^{D}$, are fully-connected layers projecting the visual and audio features, respectively, to a lower dimension $D$. The input to the transformer is ${X^e\in\mathbb{R}^{(2w+2)\times D}}$. 
In appendix~\ref{sec:ablation}, we compare the absolute positional encodings with relative \cite{shaw_2018_naacl} and Fourier feature positional encodings \cite{jaegle2021perceiver}. 

\noindent\textbf{Transformer and classifier}. We use a transformer encoder $f(\cdot)$ to process sequential audio-visual inputs, $Z=f(X^e)$. We share the weights of the transformer encoder layer-wise. 
In appendix~\ref{sec:ablation}, we compare this to a version without weight sharing.
Weight sharing uses $2.7\times$ less parameters with comparable results. 
A two-head classifier $h(\cdot)$ for verbs and nouns then predicts the sequence of $w$ actions from both the transformed visual and audio tokens $\hat{Y} =h(Z_{1:2w})$, and the action $b_i$ from the summary embeddings $\hat{y} =h(Z_{2w:2w+2})$.

\noindent \textbf{Loss function}. Recall that our goal is to classify $b_i$, the action localised at the centre of our temporal context. Nevertheless, we can leverage the ground-truth of neighbouring actions within $w$ for additional supervision to train the audio-visual transformer. Our loss is composed of two terms, the main loss for training the model to classify the action at the centre of our temporal context $i=\frac{w}{2}$, and an auxiliary loss to predict all actions in the sequence:
\vspace*{-6pt}
\begin{align}
        L_{\text{m}}&=CE(Y_i^\text{V},\hat{y}^{\text{V}})+CE(Y_i^\text{N},\hat{y}^{\text{N}}) \\
        L_{\text{a}}&=\sum_{j=1}^w \big(CE(Y_j^\text{V},\hat{Y}_j^\text{V})+CE(Y_j^\text{V},\hat{Y}_{|w|+j}^\text{V})+ CE(Y_j^\text{N},\hat{Y}_j^\text{N})+CE(Y_j^\text{N},\hat{Y}_{|w|+j}^\text{N}) \big)\\
        loss &= \beta L_{\text{m}}+(1-\beta)L_{\text{a}}\text{,}
    \label{eq:loss}
\end{align}
where $CE()$ is a cross-entropy loss, and $Y=(Y_1, ...,Y_w)$ is the ground-truth of the sequence, while $\hat{Y}_1,...,\hat{Y}_w$ and $\hat{Y}_{w+1},...,\hat{Y}_{2w}$ correspond to predictions from the transformed visual and auditory inputs respectively. 
We use $\beta$ to weight the importance of the auxiliary loss.

\subsection{Language Model}
\label{subsec:lm}

In addition to input context from the visual and audio domains, we introduce output context using a language model. 
Language modelling, commonly applied to predict the probability of a sequence of \textit{words}, is a fundamental task in NLP research~\cite{peters2018elmo,devlin2018bert, brown2020language,yang2019xlnet}. Our language model predicts the probability of a sequence of \textit{actions}. We use the language model to improve the predictions of the audio-visual transformer by filtering out improbable sequences.
 
We adopt the popular Masked Language Model (MLM), introduced in BERT~\cite{devlin2018bert}. We train this model independently from the audio-visual transformer.  
Specifically, given a sequence of actions $Y=(Y_{i-\frac{w-1}{2}}, ...,Y_{i+\frac{w-1}{2}})$, we randomly mask any action $Y_{j}$ and train the model to predict it.  For example, an input sequence to the model (for $w=5$) would be: (`turn on tap', `wash hands', <MASK>, `pick up towel', `dry hands'). Without any visual or audio input, the language model is tasked to learn a high prior probability for `turn off tap', which is masked. Note that the model is trained using the ground-truth sequence of actions.
For input representation, we split the action into verb and noun tokens (e.g. `dry hands' $\rightarrow$ `dry' and `hand'), convert them to one-hot vectors and input them into separate word-embedding layers\footnote{Learning a word embedding outperforms pretrained embeddings.}. MLM takes as input the concatenation of verb and noun embeddings. The outputs are scores for verb and noun classes using a two-head classifier, and the model is trained with a cross-entropy loss per output.

\subsection{Inference}
\label{subsec:inference}
\vspace*{-2pt}
Given the the scores of the sequence, $\hat{S}=(\hat{y}_{i-\frac{w-1}{2}},...,\hat{y}_{i+\frac{w-1}{2}})$, from the audio-visual model\footnote{These are temporally ordered predictions from the summary embeddings, and thus different from $\hat{Y}$.}, we apply a beam-search of size $K$ to find the $K$ most probable action sequences $\hat{S}_b$. Therefore, $\hat{S}_b$ is of size $K\times w$. In inference, the trained language model takes as input $\hat{S}_b$, i.e.\ it operates on sequences predicted from the audio-visual transformer.

For each sequence $l$, we calculate the probability of the sequence $p_{LM}(\hat{S}_b^l)$ from the language model by utilising the method introduced in ~\cite{shin2019effective}. We mask actions, one at a time, and predict their probability. $p_{LM}(\hat{S}_b^l)$ is the sum of log probabilities of all actions in $l$. We also calculate the probability $p_{AV}(\hat{S}_b^l)$ by summing the log probabilities of all predicted actions in $l$ by the audio-visual model. Then, we combine the probabilities of sequences of the audio-visual and language models:

\vspace*{-12pt}
\begin{equation}
  p(\hat{S}_b^l) =  \lambda p_{LM}(\hat{S}_b^l) + (1 - \lambda) p_{AV}(\hat{S}_b^l) \text{.}
  \label{eq:fusion}
\end{equation}

Sequences are then sorted in descending order by $p(\hat{S}_b^l)$. The score of the centre action from the sequence with the highest probability, is used as the final prediction.
 
\vspace*{-10pt}
\section{Experiments}
\label{sec:experiments}
\vspace*{-2pt}
\subsection{Implementation Details}
\label{subsec:dataset}

\noindent \textbf{Datasets}. EPIC-KITCHENS-100~\cite{Damen2020RESCALING} is the largest egocentric audio-visual dataset, containing unscripted daily activities, thus offering naturally variable sequences of actions. There are on average 129 actions per video (std 163 actions/video and maximum of 940 actions/video). This makes the dataset ideal for exploring temporal context. The length of sequences of $w=9$ actions (this is our default window length) is 34.4 seconds of video on average with an std of 27.8 seconds\footnote{minimum of 3.4 seconds to a maximum of 720.2 seconds.}. 
EGTEA~\cite{Li_2018_ECCV} is another video-only egocentric dataset. There are 28 hours of untrimmed cooking activities with 10K annotated action segments. Although the dataset does not contain audio, it has sequential actions annotated within long videos, and we use it to train part of our approach (vision and language).

\noindent\textbf{Visual features}. For EPIC-KITCHENS, we extract visual features with SlowFast \cite{Feichtenhofer_2019_ICCV}, using the public model and code from \cite{Damen2020RESCALING}. We first train the model with slightly different hyperparameters, where we sample a clip of 2s from an action segment,
do not freeze batch normalisation layers, and warm-up training during the first epoch starting from a learning rate of 0.001. We note that this gave us better performance. All unspecified hyperparameters remain unchanged. 
For feature extraction, 10 clips of 1s each are uniformly sampled for each action segment, with a center crop per clip. The resulting features have a dimensionality of $d_v = 2304$.
The SlowFast visual features are used for all the results in this paper, apart for the comparison with the state of the art in Table~\ref{tab:sota} where we additionally experiment with features from Mformer-HR~\cite{patrick_2021_arXiv}. These are extracted from the EPIC-KITCHENS pretrained model using a single crop per clip. The resulting features have a dimensionality of $d_v$=768.
For EGTEA, see appendix~\ref{sec:egtea_details}. 

\noindent\textbf{Auditory features}. We use Auditory SlowFast \cite{kazakos_2021_ICASSP}
for audio feature extraction when present. 
Similarly to the visual features, we extract 10 clips of 1s each uniformly spaced for each action segment, with average pooling and concatenation of the features from the Slow and Fast streams, and the resulting features have the same dimensionality, $d_a = 2304$. 

\noindent\textbf{Architectural details}. Both the audio-visual transformer encoder and the language model consist of 4 layers with shared weights, 8 attention heads and a hidden unit dimension of 512. In the audio-visual transformer, positional/modality encodings as well as verb/noun tokens have also dimensionality $D=512$ and are initialised to $\mathcal{N}(0,0.001)$. The layers $g_v(\cdot)$ and $g_a(\cdot)$ reduce the features to the common dimension $D=512$. In the encoding layer, dropout is applied at the inputs of $g_v(\cdot)$ and $g_a(\cdot)$ as well as at $X^e$. In the language model, both verb and noun word-embedding layers have a dimension of 256, and positional encodings have a dimension of 512, while dropout is applied to its inputs.

\noindent \textbf{Scheduled sampling}.
We modify the scheduled sampling from \cite{bengio2015scheduled} to train the language model. At each training iteration, we randomly mask an action, predict it, and replace the corresponding ground-truth with the prediction.

\noindent \textbf{Train / Val details}. 
For EPIC-KITCHENS, the audio-visual transformer is trained using SGD, a batch size of 32 and a learning rate of 0.01 for 100 epochs. Learning rate is decayed by a factor of 0.1 at epochs 50 and 75. In the loss function, we set $\beta=0.9$. For regularisation, a weight decay of 0.0005 is used and a dropout 0.5 and 0.1 for the encoding layers and transformer layers respectively. We use mixup data augmentation \cite{Zhang_2018_ICLR} with $\alpha=0.2$. The language model is trained for the same number of epochs with a batch size of 64, Adam optimiser with initial learning rate of 0.001 and the learning rate is decreased by a factor of 0.1 when validation accuracy saturates for over 10 epochs. The values of dropout and weight decay are the same as those of the audio-visual model. For inference, we tune $\lambda$ in Equation~\ref{eq:fusion} on the validation set with
grid-search from the set $\{0, 0.05, 0.1, 0.15, 0.2, 0.25, 0.3\}$, and we use a beam size of $K=10$. 
For training the audio-visual transformer, we randomly sample 1 out of 10 features per action. For testing, we feed all 10 features per action to the transformer and we share the positional encoding corresponding to an action with all 10 features. We also tried single feature per action followed by averaging 10 predictions but observed no difference in performance. 
For the train/val details in EGTEA, please see appendix~\ref{sec:egtea_details}.

\noindent\textbf{Evaluation metrics}. For EPIC-KITCHENS-100, we follow~\cite{Damen2020RESCALING} and report top-1 and top-5 accuracy for the validation and test sets separately (Test results and results on EPIC-KITCHENS-55 in appendix~\ref{sec:epic-kitchens-100} an \ref{sec:epic-kitchens-55}). We also follow~\cite{Damen2020RESCALING} and report results for two subsets within val/test: unseen participants and tail classes. For EGTEA, we follow  \cite{Min_2021_WACV,Lu_2019_ICCV,Kapidis_2019_ICCV} and report top-1 accuracy and mean class accuracy using the first train/test split. 

\vspace*{-10pt}
\subsection{Analysis of temporal context length}
\label{subsec:temporal_context_length}
\vspace*{-2pt}

\begin{table*}[t]
    \centering
\resizebox{\textwidth}{!}{%
\begin{tabular}{@{}lrrrrrrrrrrrr@{}}
\toprule
                                                                                     & \multicolumn{6}{c}{Overall}                           & \multicolumn{3}{c}{Unseen Participants} & \multicolumn{3}{c}{Tail-classes} \\
                                                                                      \cmidrule(r){2-7}                                       \cmidrule(lr){8-10}                   \cmidrule(l){11-13}
                                                                                     & \multicolumn{3}{c}{Top-1 Accuracy (\%)} & \multicolumn{3}{c}{Top-5 Accuracy (\%)} & \multicolumn{3}{c}{Top-1 Accuracy (\%)}           & \multicolumn{3}{c}{Top-1 Accuracy (\%)}    \\
                                                                                      \cmidrule(r){2-4}           \cmidrule(lr){5-7}          \cmidrule(lr){8-10}                   \cmidrule(l){11-13}
                                      $w$                                    & \multicolumn{1}{c}{Verb} & \multicolumn{1}{c}{Noun} & \multicolumn{1}{c}{Action} & \multicolumn{1}{c}{Verb} & \multicolumn{1}{c}{Noun} & \multicolumn{1}{c}{Action}& \multicolumn{1}{c}{Verb} & \multicolumn{1}{c}{Noun} & \multicolumn{1}{c}{Action} & \multicolumn{1}{c}{Verb} & \multicolumn{1}{c}{Noun} & \multicolumn{1}{c}{Action} \\ \midrule

            								1	& 67.93 & 52.29 & 41.30 & 90.53 & 76.47 & 61.52 & 61.13 & 44.60 & 32.58 & 42.05 & 27.42 & 21.48\\
            								3 & 69.80 & 55.24 & 43.52 & 91.30 & 79.04 & 63.25 & 61.41 & 46.48 & 33.71 & 39.09 & 32.58 & 23.06\\
            								5 & 70.38 & 56.16 & 45.13 & \textbf{91.67} & \textbf{79.47} & \textbf{64.14} & 61.97 & 46.95 & 34.74 & \textbf{43.12} & 32.53 & 24.54\\
            								7 & 70.43 & 56.19 & 45.01 & 91.23 & 79.13 & 63.52 & 62.63 & \textbf{47.14} & 34.84 & 41.31 & 32.79 & 24.12\\
            								9 & \textbf{70.60} & \textbf{56.26} & \textbf{45.48} & 91.14 & 79.06 & 63.06 & \textbf{63.76} & \textbf{47.14} & \textbf{35.87} & 41.36 & 32.84 & \textbf{24.70}\\
            								11 & 70.55 & 55.74 & 44.68 & 91.18 & 79.23 & 63.02 & 62.91 & 46.57 & 34.74 & 41.82 & \textbf{33.58} & 24.44\\
            							\bottomrule
\end{tabular}}
    \caption{Analysis of temporal context extent for MTCN on EPIC-KITCHENS.}
    \vspace*{-10pt}
    \label{tab:temporal_context_extent}
\end{table*}

We first analyse the importance of the temporal context extent by varying the size of $w$, i.e.\ the length of window of actions that our model observes. We use the validation set of EPIC-KITCHENS for this analysis as well as for the ablations in Sec.~\ref{sec:ablations}, and report EGTEA analysis in appendix~\ref{sec:egtea_ablation}. We perform this analysis both for MTCN containing all modalities as shown in Table~\ref{tab:temporal_context_extent}, as well as for the language model as shown in Table~\ref{tab:temporal_context_extent_LM}. 
Varying the length of the window has a big impact on the model's accuracy showcasing that MTCN successfully utilises temporal context. Using temporal context outperforms $w=1$, i.e.\ no temporal context. As the window length increases the performance also increases. Overall top-1 accuracy increases up to $w=9$ while top-5 up to a window $w=5$. Performance on unseen participants and tail classes also increases up to $w=9$.

\begin{table}[t]
\begin{minipage}{0.3\textwidth}
\resizebox{0.9\textwidth}{!}{%
\begin{tabular}{@{}lrrr@{}}
\toprule
                                                                                     & \multicolumn{3}{c}{Overall}                            \\
                                                                                      \cmidrule(r){2-4}                                                        & \multicolumn{3}{c}{Top-1 Accuracy (\%)} \\
                                                                                      \cmidrule(r){2-4} 
                                      $w$                                    & \multicolumn{1}{c}{Verb} & \multicolumn{1}{c}{Noun} & \multicolumn{1}{c}{Action}  \\ \midrule

                                            1												
                                & 19.32 & 3.74 & 0.82 \\
            								3														&38.08 & 45.56 & 23.85 \\
            								5														& 42.15 & \textbf{50.36} & 29.48 \\
            								7														& 42.93 & 50.35 & \textbf{29.91} \\
            								9														& \textbf{43.06} & 50.22 & 29.41 \\
            								11														& 41.89 & 49.96 & 29.14 \\
            								\bottomrule
\end{tabular}}
    \caption{Analysis of the temporal context on LM.}
    \label{tab:temporal_context_extent_LM}
\end{minipage}
\hspace{0.02\textwidth}
\begin{minipage}{0.65\textwidth}
    \includegraphics[width=0.49\textwidth]{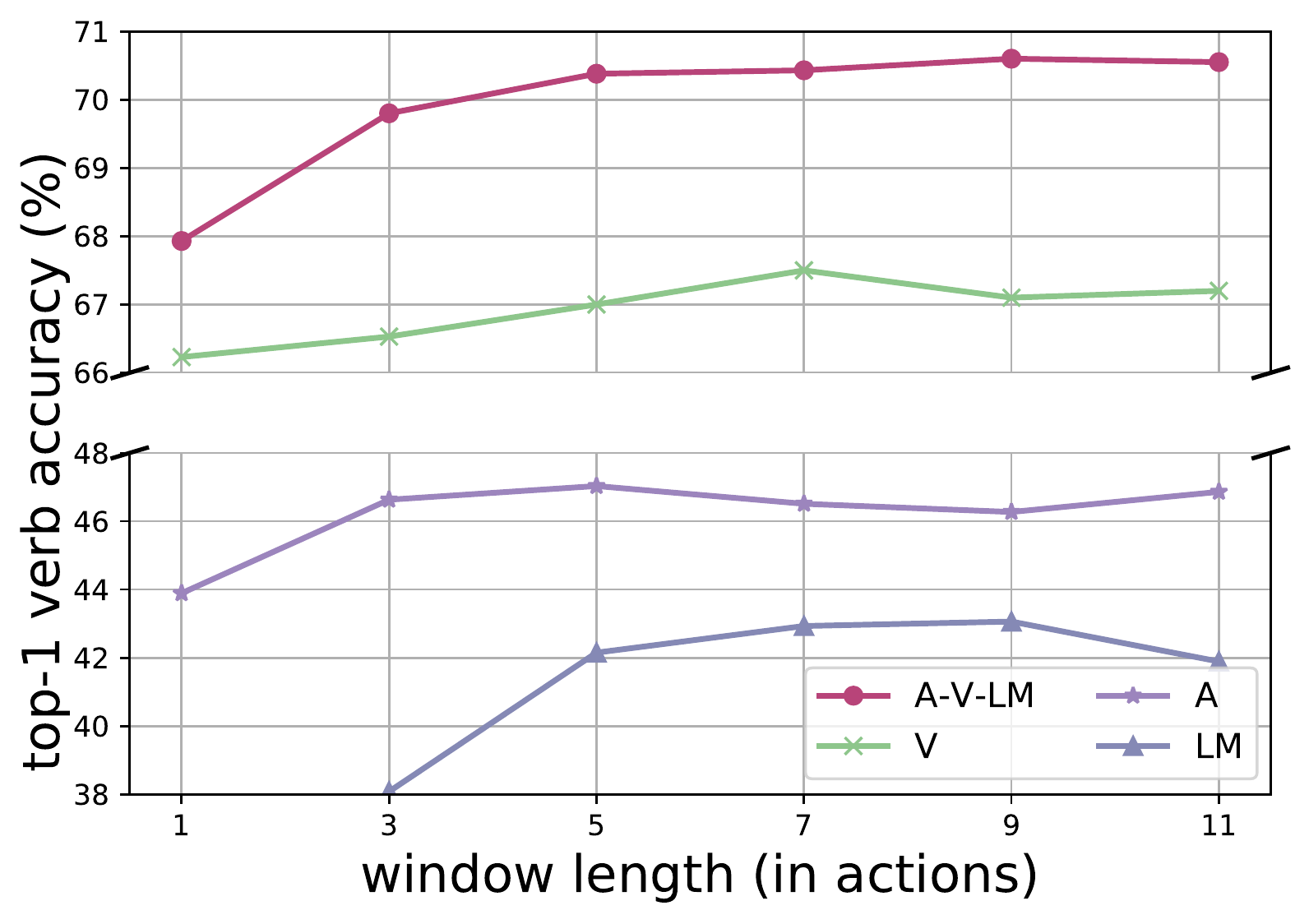}
    \includegraphics[width=0.49\textwidth]{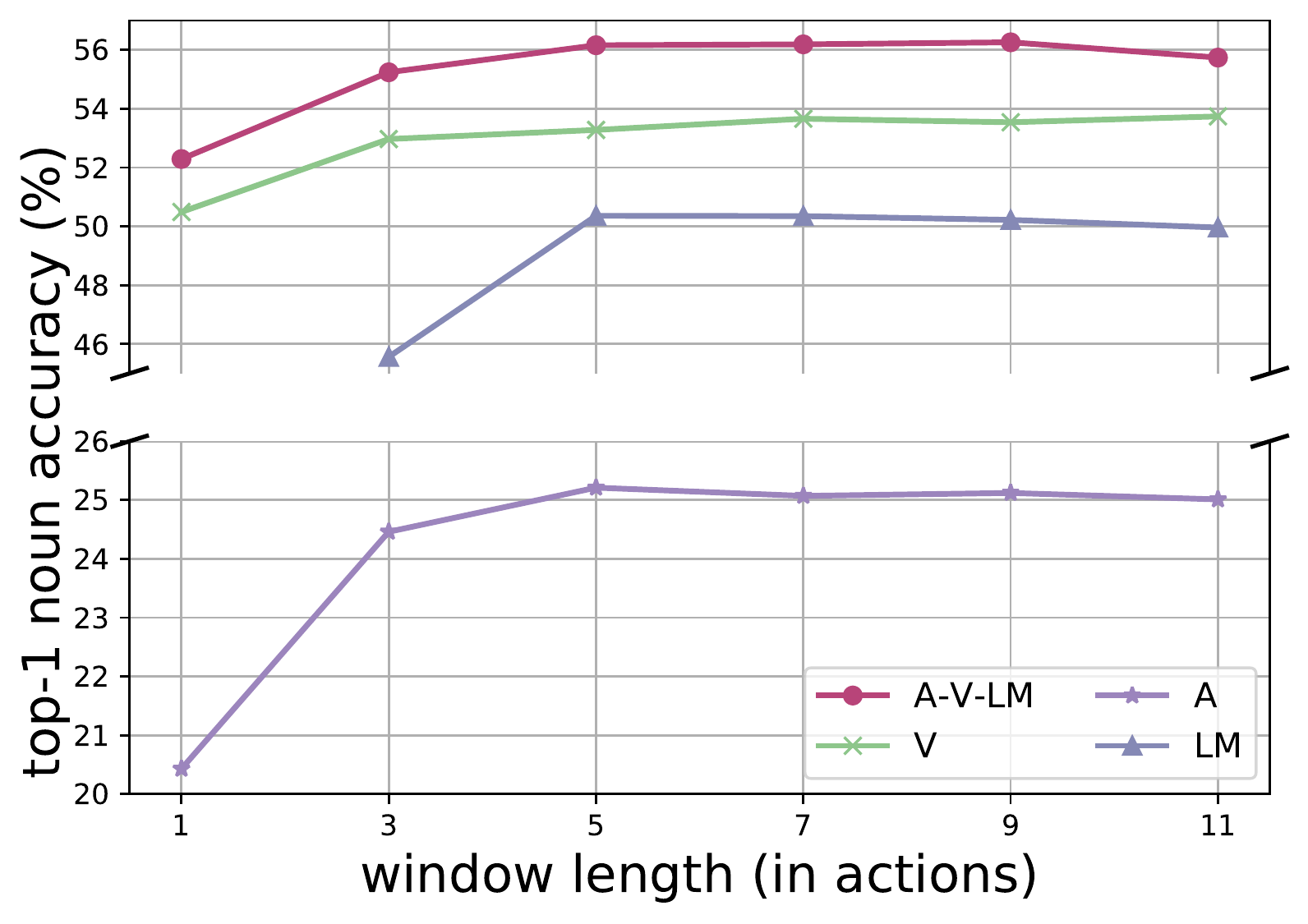}\vspace*{-12pt}
    \captionof{figure}{Effect of temporal context on verb (left) and noun (right) accuracy of individual modalities and for MTCN on EPIC-KITCHENS. Y-axis cut to emphasise details.}
    \label{fig:window_len}
    \end{minipage}
    \vspace*{-16pt}
\end{table}

In Table~\ref{tab:temporal_context_extent_LM}, experiments are conducted by masking the centre action and measuring how well the model predicts it. We use ground-truth for the other actions, as in this experiment we are interested in the maximum possible performance of the language model, i.e.\ assuming correct predictions from the audio-visual model. Here $w=1$ corresponds to a language model that randomly guesses the masked action without any context. 
When varying the length of the window for the language model, verb performance increases when enlarging the temporal context from $w=3$ to $w=9$ while for nouns optimal temporal context is $w=5$, and $w=7$ for actions. Interestingly, the language model is performing particularly well for nouns, 
since the same object is often used for the entire sequence.   
 
Finally, Fig.~\ref{fig:window_len} shows the effect of temporal context extent on verb and noun accuracy of the individual modalities as well as of MTCN. 
For verbs, visual modality performance increases up to 7 actions and then decreases while for nouns, it steadily increases up to 11 actions, possibly because larger context is able to resolve ambiguities due to occlusion. Audio can better capture temporal context for verbs than the language model, showcasing that the progression of sounds conveys more useful information about the action. For nouns, the language model outperforms audio. Moreover, the language model performs better on nouns because of repetitions of the same object in the sequence. MTCN utilising all modalities (A-V-LM) benefits the most from larger temporal context, particularly for verbs. 
With this analysis completed, we fix $w$ to 9 in all subsequent experiments.

\vspace*{-2pt}
\subsection{Results and Ablations}
\label{sec:ablations}
\vspace*{-2pt}
We compare our approach with the state-of-the-art (SOTA) approaches on EPIC-KITCHENS-100 as shown in Table~\ref{tab:sota}. MTCN significantly outperforms convolutional approaches \cite{TSN2016ECCV,lin2019tsm,kazakos2019TBN,Feichtenhofer_2019_ICCV}. We outperform the audio-visual TBN \cite{kazakos2019TBN} by 8\% on top-1 actions, and SlowFast~\cite{Feichtenhofer_2019_ICCV} by 6\% (using the same visual features). 
We also outperform very recent transformer-based approaches~\cite{Arnab_2021_arXiv,bulat_2021_arXiv,patrick_2021_arXiv,Nagrani21c}, reporting published results (only top-1 accuracy).
Note, that MTCN consists of a lightweight transformer that operates on pre-extracted features, while \cite{Arnab_2021_arXiv,bulat_2021_arXiv,patrick_2021_arXiv,Nagrani21c} are high-capacity models trained end-to-end. 

Additionally, when we employ visual features from Mformer-HR~\cite{patrick_2021_arXiv}, MTCN improves over all transformer-based approaches, including audio-visual fusion~\cite{Nagrani21c}, by 3.5\% on top-1 nouns and 5\% on actions.
We attribute the boost on nouns to the enhanced object recognition performance of the ViT backbone~\cite{dosovitskiy2020image} in Mformer-HR and its large-scale pretraining.
These results, however, further demonstrate the potential to boost other methods and features by utilising multimodal temporal context.

\begin{table}[t]
    \centering
\resizebox{\textwidth}{!}{%
\begin{tabular}{@{}lrrrrrrrrrrrr@{}}
\toprule
                                                                                     & \multicolumn{6}{c}{Overall}                           & \multicolumn{3}{c}{Unseen Participants} & \multicolumn{3}{c}{Tail-classes} \\
                                                                                      \cmidrule(r){2-7}                                       \cmidrule(lr){8-10}                   \cmidrule(l){11-13}
                                                                                     & \multicolumn{3}{c}{Top-1 Accuracy (\%)} & \multicolumn{3}{c}{Top-5 Accuracy (\%)} & \multicolumn{3}{c}{Top-1 Accuracy (\%)}           & \multicolumn{3}{c}{Top-1 Accuracy (\%)}    \\
                                                                                      \cmidrule(r){2-4}           \cmidrule(lr){5-7}          \cmidrule(lr){8-10}                   \cmidrule(l){11-13}
                                        Model                                 & \multicolumn{1}{c}{Verb} & \multicolumn{1}{c}{Noun} & \multicolumn{1}{c}{Action} & \multicolumn{1}{c}{Verb} & \multicolumn{1}{c}{Noun} & \multicolumn{1}{c}{Action}& \multicolumn{1}{c}{Verb} & \multicolumn{1}{c}{Noun} & \multicolumn{1}{c}{Action} & \multicolumn{1}{c}{Verb} & \multicolumn{1}{c}{Noun} & \multicolumn{1}{c}{Action} \\ \midrule

											TSN~\cite{TSN2016ECCV}   & 60.2 & 46.0 & 33.2 & 89.6 & 72.9 & 55.1 & 47.4 & 38.0 & 23.5 & 30.5 & 19.4 & 13.9 \\
                                           TBN~\cite{kazakos2019TBN}         & 66.0 & 47.2 & 36.7 & 90.5 & 73.8 & 57.7 & 59.4 & 38.2 & 29.5 & 39.1 & 24.8 & 19.1 \\
                                           TSM~\cite{lin2019tsm}     & 67.9 & 49.0 & 38.3 & 91.0 & 75.0 & 60.4 & 58.7 & 39.6 & 29.5 & 36.6 & 23.4 & 17.6 \\
                                           SlowFast~\cite{Feichtenhofer_2019_ICCV}  & 65.6 & 50.0 & 38.5 & 90.0 & 75.6 & 58.6 & 56.4 & 41.5 & 29.7 & 36.2 & 23.3 & 18.8 \\
                                           ViViT-L/16x2~\cite{Arnab_2021_arXiv}	 & 66.4 & 56.8 & 44.0 & - & - & - & - & - & - & - & - &-\\
                                           X-ViT (16x)~\cite{bulat_2021_arXiv} 			& 68.7 & 56.4 & 44.3 & - & - & - & - & - & - & - & - &-\\
                                           Mformer-HR~\cite{patrick_2021_arXiv} 	& 67.0 & 58.5 & 44.5 & - & - & - & - & - & - & - & - &-\\
                        
                        MBT~\cite{Nagrani21c} 	& 64.8 & 58.0 & 43.4 & - & - & - & - & - & - & - & - &-\\
                                           MTCN - v.f. SlowFast~\cite{Feichtenhofer_2019_ICCV}												& 70.6 & 56.3 & 45.5 & \textbf{91.1} & 79.1 & 63.1 & \textbf{63.8} & 47.1 & 35.9 & 41.4 & 32.8 & 24.7 \\
                                          MTCN - v.f. Mformer-HR~\cite{patrick_2021_arXiv} & \textbf{70.7} & \textbf{62.1} & \textbf{49.6} & 90.7 & \textbf{83.1} & \textbf{68.6} & 63.7 & \textbf{50.9} & \textbf{38.9} & \textbf{41.9} & \textbf{39.2} & \textbf{27.7}\\ \bottomrule
\end{tabular}}
    \caption{Comparison with SOTA on EPIC-KITCHENS-100 using two visual features~(`v.f.')}
    \label{tab:sota}
\end{table}

\begin{table}[t]
    \centering
\resizebox{\textwidth}{!}{%
\begin{tabular}{@{}ccccrrrrrrrrrrrr@{}}
\toprule
                                                                                 &&&& \multicolumn{6}{c}{Overall}                           & \multicolumn{3}{c}{Unseen Participants} & \multicolumn{3}{c}{Tail-classes} \\
                                                                                      \cmidrule(r){5-10}                                       \cmidrule(lr){11-13}                   \cmidrule(l){14-16}
                                                                                &&&& \multicolumn{3}{c}{Top-1 Accuracy (\%)} & \multicolumn{3}{c}{Top-5 Accuracy (\%)} & \multicolumn{3}{c}{Top-1 Accuracy (\%)}           & \multicolumn{3}{c}{Top-1 Accuracy (\%)}    \\
                                                                                      \cmidrule(r){5-7}           \cmidrule(lr){8-10}          \cmidrule(lr){11-13}                   \cmidrule(l){14-16}
                                      V & A & LM & Aux                                  & \multicolumn{1}{c}{Verb} & \multicolumn{1}{c}{Noun} & \multicolumn{1}{c}{Action} & \multicolumn{1}{c}{Verb} & \multicolumn{1}{c}{Noun} & \multicolumn{1}{c}{Action}& \multicolumn{1}{c}{Verb} & \multicolumn{1}{c}{Noun} & \multicolumn{1}{c}{Action} & \multicolumn{1}{c}{Verb} & \multicolumn{1}{c}{Noun} & \multicolumn{1}{c}{Action} \\ \midrule

									\cmark & \xmark & \xmark & \cmark					& 67.10 & 53.54 & 41.49 & 90.62 & 78.22 & 62.32 & 58.87 & 43.29 & 30.99 & 40.97 & 30.47 & 22.35\\
									\cmark & \xmark & \cmark & \cmark						& 67.84 & 54.08 & 42.05 & 90.63 & 78.20 & 60.68 & 59.06 & 44.23 & 31.36 & 39.77 & 31.32 & 22.38\\
									\cmark & \cmark & \xmark & \cmark			& 70.23 & 55.82 & 45.00 & 91.13 & 79.06 & \textbf{64.58} & 63.29 & 46.38 & 35.02 & \textbf{41.76} & 32.26 & 24.41\\
									\cmark & \cmark & \cmark & \xmark					& 69.31 & 55.46 & 43.81 & \textbf{91.19} & \textbf{79.76} & 62.35 & 61.13 & 46.01 & 33.90 & 39.55 & 30.74 & 22.74\\
							       	    \cmidrule{1-16}\morecmidrules\cmidrule{1-16}
									\cmark & \cmark & \cmark & \cmark                  & \textbf{70.60} & \textbf{56.26} & \textbf{45.48} & 91.14 & 79.06 & 63.06 & \textbf{63.76} & \textbf{47.14} & \textbf{35.87} & 41.36 & \textbf{32.84} & \textbf{24.70}\\
									\cmidrule{1-16}\morecmidrules\cmidrule{1-16}
            						\cmark & \cmark & $\dag$ & \cmark & 71.33 & 63.56 & 50.32 & 92.05 & 83.16 & 67.85 & 62.25 & 52.68 & 37.56 & 41.53 & 44.16 & 29.89\\\bottomrule

\end{tabular}}
    \caption{Ablation on multimodal temporal context and auxiliary loss in EPIC-KITCHENS. $\dag$: upper bound using ground-truth knowledge as input to the language model.}
    \label{tab:multimodal_context}
\end{table}

\begin{wraptable}[11]{r}{0.4\textwidth}
    \centering
\resizebox{0.4\textwidth}{!}{%
\begin{tabular}{>{\color{black}}l>{\color{black}}r>{\color{black}}r}
\toprule
                                        Method                                   & \multicolumn{1}{c}{\textcolor{black}{Top-1 (\%)}} & \multicolumn{1}{c}{\textcolor{black}{MC(\%)}}\\\midrule

										   Li et al.~\cite{LiYin_2018_ECCV}	            & - & 53.30 \\
                                           Ego-RNN~\cite{Sudhakaran_2018_BMVC}          & 62.17 & - \\
                                           Kapidis et al.~\cite{Kapidis_2019_ICCV}      & 68.99 & 61.40\\
                                           Lu et al.~\cite{Lu_2019_ICCV}                & 68.60 & 60.54\\
                                           SlowFast~\cite{Feichtenhofer_2019_ICCV} & 70.43 & 61.92\\
                                           MCN~\cite{Huang_2020_TIP}                    & 55.63 & -\\
                                           Min et al.~\cite{Min_2021_WACV}              & 69.58 & 62.84\\
                                           MTCN (V) (Ours) &72.55 &64.86\\
                                           MTCN (V+LM) (Ours)  & \textbf{73.59} & \textbf{65.87}\\\bottomrule

\end{tabular}}
    \caption{Comparative results on EGTEA. MC: Mean Class}
    \label{tab:sota_egtea}
\end{wraptable}
We compare MTCN without audio with the state-of-the-art on EGTEA, in Table~\ref{tab:sota_egtea}. This model uses $w=3$. Our model improves over previous approaches by 3\% in both top-1 and mean class accuracy. Note that MTCN outperforms SlowFast~\cite{Feichtenhofer_2019_ICCV} which we used to extract features. In 
appendix~\ref{sec:egtea_ablation}, we ablate the temporal context and language model, where we showcase that the language model provides a bigger boost in performance, possibly due to the absence of the audio modality (also shown in Table~\ref{tab:sota_egtea}: `V' vs `V+LM').

\begin{figure}[t]
    \centering
    \includegraphics[width=\textwidth]{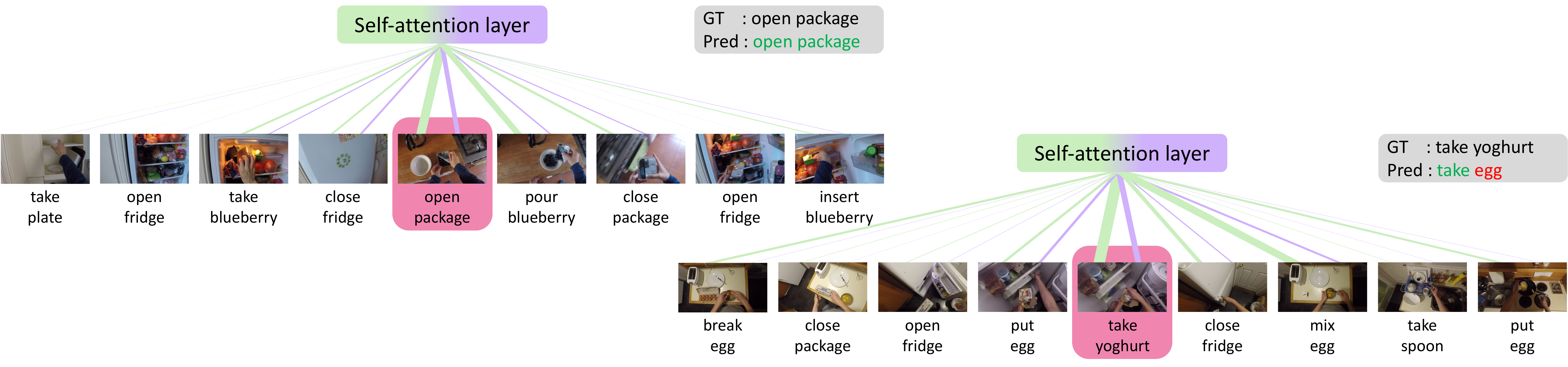}
    \vspace*{-16pt}
    \caption{Qualitative results of attention weights along with the predictions of our model in EPIC-KITCHENS. Green and purple edges represent attention from the noun embedding to visual and auditory tokens, respectively. 
    Thickness indicates attention weights magnitude to centre (bordered) and temporal context actions.} 
    \label{fig:att_visualisation}
     \vspace*{-12pt}
\end{figure}


\noindent\textbf{Multimodality Ablation}. 
We offer an ablation to identify the performance impact of the components of our MTCN. Results are shown in Table~\ref{tab:multimodal_context}. 
We remove audio and language modalities from our model's input and output respectively to assess their importance. Multimodal context is important, as our proposed model enjoys considerable margins compared with the model trained only with visual context (line 1 in Table~\ref{tab:multimodal_context}). Audio is beneficial, confirming the finding of prior works. 

Although the language model provides smaller boost in performance than audio, it showcases that it is useful to model prior temporal context at the output level, and that its benefits are complementary to audio.
We also include upper bound performance improvement from the addition of the language model, where it takes as input the ground-truth preceding/succeeding actions rather than the predictions from the audio-visual transformer (last line in Table~\ref{tab:multimodal_context}), effectively a language infilling problem. These results demonstrate the potential margin for improvement, particularly for nouns and accordingly action accuracy as well as tail classes.
In appendix~\ref{sec:lm_analysis}, we report statistical significance over multiple runs for the language model, and compare it to alternative baselines. 

\noindent\textbf{Auxiliary loss}. Also from Table~\ref{tab:multimodal_context}, training MTCN with auxiliary loss boosts its performance almost in all metrics, confirming that utilising supervision from neighbouring actions can improve the performance of the action of interest, i.e. the action at the centre of the window. 

While we focus on offline action recognition, we 
can evaluate our model for online recognition by predicting the last action in the sequence instead of the centre one. Results are included in appendix~\ref{sec:online} for different values of $w$, where $w=7$ provided best results for EPIC-KITCHENS. Best top-1 action performance using past context solely is 42.96\% compared to 45.48\% using surrounding context. However, we demonstrate that our model can leverage multimodal temporal context in this setting.

In Fig.~\ref{fig:att_visualisation}, we visualise the auditory and visual temporal context attention on a correctly recognised sequence (left) -- where subsequent actions of pouring and closing are particularly informative, and an incorrectly (right) predicted sequence -- where both attention weights are high on actions containing the egg, causing the model to incorrectly  predict the noun as egg. More qualitative examples are included in appendix~\ref{sec:attention}.

\vspace*{-12pt}
\section{Conclusion}
\label{sec:conclusion}
\vspace*{-2pt}
We formulate temporal context as a sequence of actions, and utilise past and future context to enhance the prediction of the centre action in the sequence. We propose MTCN, a model that attends to vision and audio as input modality context, and language as output modality context. We train MTCN with additional supervision from neighbouring actions. Our results showcase the importance of audio and language as additional modality context. We report SOTA results on two egocentric video datasets: EPIC-KITCHENS and EGTEA. An extension to MTCN would incorporate an actionness score of neighbouring frames, to distinguish background frames and learn from action sequences in untrimmed videos without temporal bounds during testing. This would bridge the problems of recognition and detection, utilising multimodality and temporal context. We will be exploring this in future work. 
\vspace*{-12pt}
\paragraph{Acknowledgements.} This work used public datasets and is supported by EPSRC Doctoral Training Program, EPSRC UMPIRE (EP/T004991/1) and the EPSRC Programme Grant VisualAI (EP/T028572/1). Jaesung Huh is funded by a Global Korea Scholarship.
\bibliography{egbib}

\begin{thebibliography}{76}
\providecommand{\natexlab}[1]{#1}
\providecommand{\url}[1]{\texttt{#1}}
\expandafter\ifx\csname urlstyle\endcsname\relax
  \providecommand{\doi}[1]{doi: #1}\else
  \providecommand{\doi}{doi: \begingroup \urlstyle{rm}\Url}\fi

\bibitem[Arnab et~al.(2021)Arnab, Dehghani, Heigold, Sun, Lucic, and
  Schmid]{Arnab_2021_arXiv}
Anurag Arnab, Mostafa Dehghani, Georg Heigold, Chen Sun, Mario Lucic, and
  Cordelia Schmid.
\newblock Vivit: A video vision transformer.
\newblock In \emph{Proceedings of International Conference on Computer Vision
  (ICCV)}, 2021.

\bibitem[Baradel et~al.(2018)Baradel, Neverova, Wolf, Mille, and
  Mori]{Baradel_2018_ECCV}
Fabien Baradel, Natalia Neverova, Christian Wolf, Julien Mille, and Greg Mori.
\newblock Object level visual reasoning in videos.
\newblock In \emph{Proceedings of European Conference on Computer Vision
  (ECCV)}, 2018.

\bibitem[Beery et~al.(2020)Beery, Wu, Rathod, Votel, and
  Huang]{Beery_2020_CVPR}
Sara Beery, Guanhang Wu, Vivek Rathod, Ronny Votel, and Jonathan Huang.
\newblock Context r-cnn: Long term temporal context for per-camera object
  detection.
\newblock In \emph{Proceedings of Conference on Computer Vision and Pattern
  Recognition (CVPR)}, 2020.

\bibitem[Bengio et~al.(2015)Bengio, Vinyals, Jaitly, and
  Shazeer]{bengio2015scheduled}
Samy Bengio, Oriol Vinyals, Navdeep Jaitly, and Noam Shazeer.
\newblock Scheduled sampling for sequence prediction with recurrent neural
  networks.
\newblock In \emph{Advances in Neural Information Processing Systems
  (NeurIPS)}, 2015.

\bibitem[Bengio et~al.(2003)Bengio, Ducharme, Vincent, and
  Janvin]{bengio2003neural}
Yoshua Bengio, R{\'e}jean Ducharme, Pascal Vincent, and Christian Janvin.
\newblock A neural probabilistic language model.
\newblock \emph{Journal of Machine Learning Research (JMLR)}, 3:\penalty0
  1137--1155, 2003.

\bibitem[Bertasius and Torresani(2020)]{Bertasius_2020_CVPR}
Gedas Bertasius and Lorenzo Torresani.
\newblock Classifying, segmenting, and tracking object instances in video with
  mask propagation.
\newblock In \emph{Proceedings of Conference on Computer Vision and Pattern
  Recognition (CVPR)}, 2020.

\bibitem[Bertasius et~al.(2021)Bertasius, Wang, and
  Torresani]{Bertasius_2021_arXiv}
Gedas Bertasius, Heng Wang, and Lorenzo Torresani.
\newblock Is space-time attention all you need for video understanding?
\newblock In \emph{Proceedings of International Conference on Machine Learning
  (ICML)}, 2021.

\bibitem[Brown et~al.(2020)Brown, Mann, Ryder, Subbiah, Kaplan, Dhariwal,
  Neelakantan, Shyam, Sastry, Askell, Agarwal, Herbert-Voss, Krueger, Henighan,
  Child, Ramesh, Ziegler, Wu, Winter, Hesse, Chen, Sigler, Litwin, Gray, Chess,
  Clark, Berner, McCandlish, Radford, Sutskever, and Amodei]{brown2020language}
Tom Brown, Benjamin Mann, Nick Ryder, Melanie Subbiah, Jared~D Kaplan, Prafulla
  Dhariwal, Arvind Neelakantan, Pranav Shyam, Girish Sastry, Amanda Askell,
  Sandhini Agarwal, Ariel Herbert-Voss, Gretchen Krueger, Tom Henighan, Rewon
  Child, Aditya Ramesh, Daniel Ziegler, Jeffrey Wu, Clemens Winter, Chris
  Hesse, Mark Chen, Eric Sigler, Mateusz Litwin, Scott Gray, Benjamin Chess,
  Jack Clark, Christopher Berner, Sam McCandlish, Alec Radford, Ilya Sutskever,
  and Dario Amodei.
\newblock Language models are few-shot learners.
\newblock In \emph{Advances in Neural Information Processing Systems
  (NeurIPS)}, 2020.

\bibitem[Bulat et~al.(2021)Bulat, Perez{-}Rua, Sudhakaran, Mart{\'{\i}}nez, and
  Tzimiropoulos]{bulat_2021_arXiv}
Adrian Bulat, Juan{-}Manuel Perez{-}Rua, Swathikiran Sudhakaran, Brais
  Mart{\'{\i}}nez, and Georgios Tzimiropoulos.
\newblock Space-time mixing attention for video transformer.
\newblock In \emph{Advances in Neural Information Processing Systems
  (NeurIPS)}, 2021.

\bibitem[Cartas et~al.(2021)Cartas, Radeva, and Dimiccoli]{Cartas_2021_ICPR}
Alejandro Cartas, Petia Radeva, and Mariella Dimiccoli.
\newblock Modeling long-term interactions to enhance action recognition.
\newblock In \emph{Proceedings of International Conference on Pattern
  Recognition (ICPR)}, 2021.

\bibitem[Chan et~al.(2016)Chan, Jaitly, Le, and Vinyals]{chan2016listen}
William Chan, Navdeep Jaitly, Quoc Le, and Oriol Vinyals.
\newblock Listen, attend and spell: A neural network for large vocabulary
  conversational speech recognition.
\newblock In \emph{Proceedings of the IEEE International Conference on
  Acoustics, Speech and Signal Processing (ICASSP)}, 2016.

\bibitem[Damen et~al.(2018)Damen, Doughty, Farinella, Fidler, Furnari, Kazakos,
  Moltisanti, Munro, Perrett, Price, and Wray]{Damen2018EPICKITCHENS}
Dima Damen, Hazel Doughty, Giovanni~Maria Farinella, Sanja Fidler, Antonino
  Furnari, Evangelos Kazakos, Davide Moltisanti, Jonathan Munro, Toby Perrett,
  Will Price, and Michael Wray.
\newblock Scaling egocentric vision: The epic-kitchens dataset.
\newblock In \emph{European Conference on Computer Vision (ECCV)}, 2018.

\bibitem[Damen et~al.(2021)Damen, Doughty, Farinella, , Furnari, Ma, Kazakos,
  Moltisanti, Munro, Perrett, Price, and Wray]{Damen2020RESCALING}
Dima Damen, Hazel Doughty, Giovanni~Maria Farinella, , Antonino Furnari, Jian
  Ma, Evangelos Kazakos, Davide Moltisanti, Jonathan Munro, Toby Perrett, Will
  Price, and Michael Wray.
\newblock Rescaling egocentric vision: Collection pipeline and challenges for
  epic-kitchens-100.
\newblock \emph{International Journal of Computer Vision (IJCV)}, 2021.

\bibitem[Devlin et~al.(2019)Devlin, Chang, Lee, and Toutanova]{devlin2018bert}
Jacob Devlin, Ming-Wei Chang, Kenton Lee, and Kristina Toutanova.
\newblock Bert: Pre-training of deep bidirectional transformers for language
  understanding.
\newblock In \emph{Proceedings of North American Chapter of the Association for
  Computational Linguistics (NAACL)}, 2019.

\bibitem[Dosovitskiy et~al.(2021)Dosovitskiy, Beyer, Kolesnikov, Weissenborn,
  Zhai, Unterthiner, Dehghani, Minderer, Heigold, Gelly, Uszkoreit, and
  Houlsby]{dosovitskiy2020image}
Alexey Dosovitskiy, Lucas Beyer, Alexander Kolesnikov, Dirk Weissenborn,
  Xiaohua Zhai, Thomas Unterthiner, Mostafa Dehghani, Matthias Minderer, Georg
  Heigold, Sylvain Gelly, Jakob Uszkoreit, and Neil Houlsby.
\newblock An image is worth 16x16 words: Transformers for image recognition at
  scale.
\newblock In \emph{Proceedings of International Conference on Learning
  Representations (ICLR)}, 2021.

\bibitem[Doval and G{\'o}mez-Rodr{\'\i}guez(2019)]{doval2019comparing}
Yerai Doval and Carlos G{\'o}mez-Rodr{\'\i}guez.
\newblock Comparing neural-and n-gram-based language models for word
  segmentation.
\newblock \emph{Journal of the Association for Information Science and
  Technology}, 70:\penalty0 187--197, 2019.

\bibitem[Ephrat et~al.(2018)Ephrat, Mosseri, Lang, Dekel, Wilson, Hassidim,
  Freeman, and Rubinstein]{ephrat2018looking}
Ariel Ephrat, Inbar Mosseri, Oran Lang, Tali Dekel, Kevin Wilson, Avinatan
  Hassidim, William~T Freeman, and Michael Rubinstein.
\newblock Looking to listen at the cocktail party: a speaker-independent
  audio-visual model for speech separation.
\newblock \emph{ACM Transactions on Graphics (TOG)}, 37:\penalty0 1--11, 2018.

\bibitem[Feichtenhofer et~al.(2019)Feichtenhofer, Fan, Malik, and
  He]{Feichtenhofer_2019_ICCV}
Christoph Feichtenhofer, Haoqi Fan, Jitendra Malik, and Kaiming He.
\newblock Slow{F}ast {N}etworks for {V}ideo {R}ecognition.
\newblock In \emph{Proceedings of International Conference on Computer Vision
  (ICCV)}, 2019.

\bibitem[Furnari and Farinella(2019)]{Furnari_2019_ICCV}
Antonino Furnari and Giovanni~Maria Farinella.
\newblock What would you expect? anticipating egocentric actions with
  {R}olling-{U}nrolling {LSTM}s and modality attention.
\newblock In \emph{Proceedings of International Conference on Computer Vision
  (ICCV)}, 2019.

\bibitem[Gabeur et~al.(2020)Gabeur, Sun, Alahari, and Schmid]{gabeur2020multi}
Valentin Gabeur, Chen Sun, Karteek Alahari, and Cordelia Schmid.
\newblock Multi-modal transformer for video retrieval.
\newblock In \emph{Proceedings of European Conference on Computer Vision
  (ECCV)}, 2020.

\bibitem[Gan et~al.(2020)Gan, Huang, Chen, Tenenbaum, and
  Torralba]{gan2020foley}
Chuang Gan, Deng Huang, Peihao Chen, Joshua~B Tenenbaum, and Antonio Torralba.
\newblock Foley music: Learning to generate music from videos.
\newblock In \emph{Proceedings of European Conference on Computer Vision
  (ECCV)}, 2020.

\bibitem[Girdhar et~al.(2019)Girdhar, Carreira, Doersch, and
  Zisserman]{Girdhar_2019_CVPR}
Rohit Girdhar, Joao Carreira, Carl Doersch, and Andrew Zisserman.
\newblock Video action transformer network.
\newblock In \emph{Proceedings of Conference on Computer Vision and Pattern
  Recognition (CVPR)}, 2019.

\bibitem[Graves et~al.(2006)Graves, Fern{\'a}ndez, Gomez, and
  Schmidhuber]{graves2006connectionist}
Alex Graves, Santiago Fern{\'a}ndez, Faustino Gomez, and J{\"u}rgen
  Schmidhuber.
\newblock Connectionist temporal classification: labelling unsegmented sequence
  data with recurrent neural networks.
\newblock In \emph{Proceedings of International Conference on Machine Learning
  (ICML)}, 2006.

\bibitem[Gulcehre et~al.(2017)Gulcehre, Firat, Xu, Cho, and
  Bengio]{gulcehre2017integrating}
Caglar Gulcehre, Orhan Firat, Kelvin Xu, Kyunghyun Cho, and Yoshua Bengio.
\newblock On integrating a language model into neural machine translation.
\newblock \emph{Computer Speech \& Language}, 45:\penalty0 137--148, 2017.

\bibitem[Hannun et~al.(2014)Hannun, Case, Casper, Catanzaro, Diamos, Elsen,
  Prenger, Satheesh, Sengupta, Coates, and Ng]{hannun2014deep}
Awni~Y. Hannun, Carl Case, Jared Casper, Bryan Catanzaro, Greg Diamos, Erich
  Elsen, Ryan Prenger, Sanjeev Satheesh, Shubho Sengupta, Adam Coates, and
  Andrew~Y. Ng.
\newblock Deep speech: Scaling up end-to-end speech recognition.
\newblock \emph{CoRR}, abs/1412.5567, 2014.

\bibitem[Harwath et~al.(2016)Harwath, Torralba, and
  Glass]{harwath2017unsupervised}
David Harwath, Antonio Torralba, and James Glass.
\newblock Unsupervised learning of spoken language with visual context.
\newblock In \emph{Advances in Neural Information Processing Systems
  (NeurIPS)}, 2016.

\bibitem[Huang et~al.(2020)Huang, Cai, Li, Lu, and Sato]{Huang_2020_TIP}
Yifei Huang, Minjie Cai, Zhenqiang Li, Feng Lu, and Yoichi Sato.
\newblock Mutual context network for jointly estimating egocentric gaze and
  action.
\newblock \emph{IEEE Transactions on Image Processing (TIP)}, 29:\penalty0
  7795--7806, 2020.

\bibitem[Iashin and Rahtu(2020)]{iashin2020multi}
Vladimir Iashin and Esa Rahtu.
\newblock Multi-modal dense video captioning.
\newblock In \emph{Proceedings of Conference on Computer Vision and Pattern
  Recognition Workshops (CVPRW)}, 2020.

\bibitem[Jaegle et~al.(2021)Jaegle, Gimeno, Brock, Zisserman, Vinyals, and
  Carreira]{jaegle2021perceiver}
Andrew Jaegle, Felix Gimeno, Andrew Brock, Andrew Zisserman, Oriol Vinyals, and
  Jo{\~{a}}o Carreira.
\newblock Perceiver: General perception with iterative attention.
\newblock \emph{CoRR}, abs/2103.03206, 2021.

\bibitem[Kapidis et~al.(2019)Kapidis, Poppe, van Dam, Noldus, and
  Veltkamp]{Kapidis_2019_ICCV}
Georgios Kapidis, Ronald Poppe, Elsbeth van Dam, Lucas Noldus, and Remco
  Veltkamp.
\newblock Multitask learning to improve egocentric action recognition.
\newblock In \emph{Proceedings of International Conference on Computer Vision
  Workshops (ICCVW)}, 2019.

\bibitem[Kay et~al.(2017)Kay, Carreira, Simonyan, Zhang, Hillier,
  Vijayanarasimhan, Viola, Green, Back, Natsev, Suleyman, and
  Zisserman]{kay2017kinetics}
Will Kay, Jo{\~{a}}o Carreira, Karen Simonyan, Brian Zhang, Chloe Hillier,
  Sudheendra Vijayanarasimhan, Fabio Viola, Tim Green, Trevor Back, Paul
  Natsev, Mustafa Suleyman, and Andrew Zisserman.
\newblock The kinetics human action video dataset.
\newblock \emph{CoRR}, abs/1705.06950, 2017.

\bibitem[Kazakos et~al.(2019)Kazakos, Nagrani, Zisserman, and
  Damen]{kazakos2019TBN}
Evangelos Kazakos, Arsha Nagrani, Andrew Zisserman, and Dima Damen.
\newblock Epic-fusion: Audio-visual temporal binding for egocentric action
  recognition.
\newblock In \emph{Proceedings of International Conference on Computer Vision
  (ICCV)}, 2019.

\bibitem[Kazakos et~al.(2021)Kazakos, Nagrani, Zisserman, and
  Damen]{kazakos_2021_ICASSP}
Evangelos Kazakos, Arsha Nagrani, Andrew Zisserman, and Dima Damen.
\newblock Slow-fast auditory streams for audio recognition.
\newblock In \emph{Proceedings of the IEEE International Conference on
  Acoustics, Speech and Signal Processing (ICASSP)}, 2021.

\bibitem[Kim et~al.(2016)Kim, Jernite, Sontag, and Rush]{kim2016character}
Yoon Kim, Yacine Jernite, David Sontag, and Alexander Rush.
\newblock Character-aware neural language models.
\newblock In \emph{Proceedings of AAAI Conference on Artificial Intelligence},
  2016.

\bibitem[Lee et~al.(2021)Lee, Yu, Kim, Breuel, Kautz, and
  Song]{lee2020parameter}
Sangho Lee, Youngjae Yu, Gunhee Kim, Thomas Breuel, Jan Kautz, and Yale Song.
\newblock {Parameter Efficient Multimodal Transformers for Video Representation
  Learning}.
\newblock In \emph{Proceedings of International Conference on Learning
  Representations ({ICLR})}, 2021.

\bibitem[Li et~al.(2019)Li, Zhu, Liu, and Yang]{li2019entangled}
Guang Li, Linchao Zhu, Ping Liu, and Yi~Yang.
\newblock Entangled transformer for image captioning.
\newblock In \emph{Proceedings of International Conference on Computer Vision
  (ICCV)}, 2019.

\bibitem[Li et~al.(2020)Li, Yin, Chu, Zhou, Wang, Fidler, and
  Li]{li2020learning}
Jiaman Li, Yihang Yin, Hang Chu, Yi~Zhou, Tingwu Wang, Sanja Fidler, and Hao
  Li.
\newblock Learning to generate diverse dance motions with transformer.
\newblock \emph{CoRR}, abs/2008.08171, 2020.

\bibitem[Li et~al.(2021{\natexlab{a}})Li, Yang, Ross, and
  Kanazawa]{li2021learn}
Ruilong Li, Shan Yang, David~A. Ross, and Angjoo Kanazawa.
\newblock Ai choreographer: Music conditioned 3d dance generation with aist++.
\newblock In \emph{Proceedings of International Conference on Computer Vision
  (ICCV)}, 2021{\natexlab{a}}.

\bibitem[Li et~al.(2021{\natexlab{b}})Li, Nagarajan, Xiong, and
  Grauman]{Li_2021_CVPR}
Yanghao Li, Tushar Nagarajan, Bo~Xiong, and Kristen Grauman.
\newblock Ego-exo: Transferring visual representations from third-person to
  first-person videos.
\newblock In \emph{Proceedings of Conference on Computer Vision and Pattern
  Recognition (CVPR)}, 2021{\natexlab{b}}.

\bibitem[Li et~al.(2018{\natexlab{a}})Li, Liu, and Rehg]{LiYin_2018_ECCV}
Yin Li, Miao Liu, and James~M. Rehg.
\newblock In the eye of beholder: Joint learning of gaze and actions in first
  person video.
\newblock In \emph{Proceedings of European Conference on Computer Vision
  (ECCV)}, 2018{\natexlab{a}}.

\bibitem[Li et~al.(2018{\natexlab{b}})Li, Li, and Vasconcelos]{Li_2018_ECCV}
Yingwei Li, Yi~Li, and Nuno Vasconcelos.
\newblock Resound: Towards action recognition without representation bias.
\newblock In \emph{Proceedings of European Conference on Computer Vision
  (ECCV)}, 2018{\natexlab{b}}.

\bibitem[Lin et~al.(2019)Lin, Gan, and Han]{lin2019tsm}
Ji~Lin, Chuang Gan, and Song Han.
\newblock {TSM}: Temporal shift module for efficient video understanding.
\newblock In \emph{Proceedings of International Conference on Computer Vision
  (ICCV)}, 2019.

\bibitem[Lin et~al.(2017)Lin, Inoue, and Shinoda]{lin2017ctc}
Mengxi Lin, Nakamasa Inoue, and Koichi Shinoda.
\newblock Ctc network with statistical language modeling for action sequence
  recognition in videos.
\newblock In \emph{Proceedings of the on Thematic Workshops of ACM Multimedia},
  2017.

\bibitem[Lu et~al.(2019{\natexlab{a}})Lu, Batra, Parikh, and
  Lee]{lu2019vilbert}
Jiasen Lu, Dhruv Batra, Devi Parikh, and Stefan Lee.
\newblock Vilbert: Pretraining task-agnostic visiolinguistic representations
  for vision-and-language tasks.
\newblock In \emph{Advances in Neural Information Processing Systems
  (NeurIPS)}, 2019{\natexlab{a}}.

\bibitem[Lu et~al.(2019{\natexlab{b}})Lu, Liao, and Li]{Lu_2019_ICCV}
Minlong Lu, Danping Liao, and Ze-Nian Li.
\newblock Learning spatiotemporal attention for egocentric action recognition.
\newblock In \emph{Proceedings of International Conference on Computer Vision
  Workshops (ICCVW)}, 2019{\natexlab{b}}.

\bibitem[Mikolov et~al.(2010)Mikolov, Karafi{\'a}t, Burget, {\v{C}}ernock{\`y},
  and Khudanpur]{mikolov2010recurrent}
Tom{\'a}{\v{s}} Mikolov, Martin Karafi{\'a}t, Luk{\'a}{\v{s}} Burget, Jan
  {\v{C}}ernock{\`y}, and Sanjeev Khudanpur.
\newblock Recurrent neural network based language model.
\newblock In \emph{Proceedings of Annual Conference of the International Speech
  Communication Association (INTERSPEECH)}, 2010.

\bibitem[Min and Corso(2021)]{Min_2021_WACV}
Kyle Min and Jason~J. Corso.
\newblock Integrating human gaze into attention for egocentric activity
  recognition.
\newblock In \emph{Proceedings of Winter Conference on Applications of Computer
  Vision (WACV)}, 2021.

\bibitem[Munro and Damen(2020)]{Munro_2020_CVPR}
Jonathan Munro and Dima Damen.
\newblock Multi-modal domain adaptation for fine-grained action recognition.
\newblock In \emph{Proceedings of Conference on Computer Vision and Pattern
  Recognition (CVPR)}, 2020.

\bibitem[Nagrani et~al.(2021)Nagrani, Yang, Arnab, Schmid, and Sun]{Nagrani21c}
Arsha Nagrani, Shan Yang, Anurag Arnab, Cordelia Schmid, and Chen Sun.
\newblock Attention bottlenecks for multimodal fusion.
\newblock In \emph{Advances in Neural Information Processing Systems
  (NeurIPS)}, 2021.

\bibitem[Neimark et~al.(2021)Neimark, Bar, Zohar, and
  Asselmann]{Neimark_2021_arXiv}
Daniel Neimark, Omri Bar, Maya Zohar, and Dotan Asselmann.
\newblock Video transformer network.
\newblock \emph{CoRR}, abs/2102.00719, 2021.

\bibitem[Ng and Fernando(2019)]{Ng_2019_arXiv}
Yan~Bin Ng and Basura Fernando.
\newblock Human action sequence classification.
\newblock \emph{CoRR}, abs/1910.02602, 2019.

\bibitem[Patrick et~al.(2021)Patrick, Campbell, Asano, Metze, Feichtenhofer,
  Vedaldi, and Henriques]{patrick_2021_arXiv}
Mandela Patrick, Dylan Campbell, Yuki~M. Asano, Ishan Misra~Florian Metze,
  Christoph Feichtenhofer, Andrea Vedaldi, and Jo{\~{a}}o~F. Henriques.
\newblock Keeping your eye on the ball: Trajectory attention in video
  transformers.
\newblock In \emph{Advances in Neural Information Processing Systems
  (NeurIPS)}, 2021.

\bibitem[Peters et~al.(2018)Peters, Neumann, Iyyer, Gardner, Clark, Lee, and
  Zettlemoyer]{peters2018elmo}
Matthew~E. Peters, Mark Neumann, Mohit Iyyer, Matt Gardner, Christopher Clark,
  Kenton Lee, and Luke Zettlemoyer.
\newblock Deep contextualized word representations.
\newblock In \emph{Proceedings of Conference of the North {A}merican Chapter of
  the Association for Computational Linguistics (NAACL)}, 2018.

\bibitem[Richard and Gall(2016)]{richard2016temporal}
Alexander Richard and Juergen Gall.
\newblock Temporal action detection using a statistical language model.
\newblock In \emph{Proceedings of Conference on Computer Vision and Pattern
  Recognition (CVPR)}, 2016.

\bibitem[Sener et~al.(2020)Sener, Singhania, and Yao]{Sener_2020_ECCV}
Fadime Sener, Dipika Singhania, and Angela Yao.
\newblock Temporal aggregate representations for long-range video
  understanding.
\newblock In \emph{Proceedings of European Conference on Computer Vision
  (ECCV)}, 2020.

\bibitem[Shaw et~al.(2018)Shaw, Uszkoreit, and Vaswani]{shaw_2018_naacl}
Peter Shaw, Jakob Uszkoreit, and Ashish Vaswani.
\newblock Self-attention with relative position representations.
\newblock In \emph{Proceedings of Conference of the North {A}merican Chapter of
  the Association for Computational Linguistics (NAACL)}, 2018.

\bibitem[Shin et~al.(2019)Shin, Lee, and Jung]{shin2019effective}
Joonbo Shin, Yoonhyung Lee, and Kyomin Jung.
\newblock Effective sentence scoring method using bert for speech recognition.
\newblock In \emph{Proceedings of Asian Conference on Machine Learning (ACML)},
  2019.

\bibitem[Sudhakaran and Lanz(2018)]{Sudhakaran_2018_BMVC}
Swathikiran Sudhakaran and Oswald Lanz.
\newblock Attention is all we need: Nailing down object-centric attention for
  egocentric activity recognition.
\newblock In \emph{Proceedings of British Machine Vision Conference (BMVC)},
  2018.

\bibitem[Sudhakaran et~al.(2019)Sudhakaran, Escalera, and
  Lanz]{Sudhakaran_2019_CVPR}
Swathikiran Sudhakaran, Sergio Escalera, and Oswald Lanz.
\newblock Lsta: Long short-term attention for egocentric action recognition.
\newblock In \emph{Proceedings of Conference on Computer Vision and Pattern
  Recognition (CVPR)}, 2019.

\bibitem[Sun et~al.(2019{\natexlab{a}})Sun, Baradel, Murphy, and
  Schmid]{sun2019learning}
Chen Sun, Fabien Baradel, Kevin Murphy, and Cordelia Schmid.
\newblock Learning video representations using contrastive bidirectional
  transformer.
\newblock \emph{CoRR}, abs/1906.05743, 2019{\natexlab{a}}.

\bibitem[Sun et~al.(2019{\natexlab{b}})Sun, Myers, Vondrick, Murphy, and
  Schmid]{sun2019videobert}
Chen Sun, Austin Myers, Carl Vondrick, Kevin Murphy, and Cordelia Schmid.
\newblock Videobert: A joint model for video and language representation
  learning.
\newblock In \emph{Proceedings of International Conference on Computer Vision
  (ICCV)}, 2019{\natexlab{b}}.

\bibitem[Tran et~al.(2018)Tran, Wang, Torresani, Ray, LeCun, and
  Paluri]{Tran_2018_CVPR}
Du~Tran, Heng Wang, Lorenzo Torresani, Jamie Ray, Yann LeCun, and Manohar
  Paluri.
\newblock A closer look at spatiotemporal convolutions for action recognition.
\newblock In \emph{Proceedings of Conference on Computer Vision and Pattern
  Recognition (CVPR)}, 2018.

\bibitem[Tzinis et~al.(2021)Tzinis, Wisdom, Jansen, Hershey, Remez, Ellis, and
  Hershey]{tzinis2020into}
Efthymios Tzinis, Scott Wisdom, Aren Jansen, Shawn Hershey, Tal Remez,
  Daniel~PW Ellis, and John~R Hershey.
\newblock Into the wild with audioscope: Unsupervised audio-visual separation
  of on-screen sounds.
\newblock In \emph{Proceedings of International Conference on Learning
  Representations (ICLR)}, 2021.

\bibitem[Wang et~al.(2016{\natexlab{a}})Wang, Xiong, Wang, Qiao, Lin, Tang, and
  {Van Gool}]{TSN2016ECCV}
Limin Wang, Yuanjun Xiong, Zhe Wang, Yu~Qiao, Dahua Lin, Xiaoou Tang, and Luc
  {Van Gool}.
\newblock Temporal segment networks: Towards good practices for deep action
  recognition.
\newblock In \emph{Proceedings of European Conference on Computer Vision
  (ECCV)}, 2016{\natexlab{a}}.

\bibitem[Wang et~al.(2020{\natexlab{a}})Wang, Tran, and
  Feiszli]{Wang_2020_CVPR}
Weiyao Wang, Du~Tran, and Matt Feiszli.
\newblock What makes training multi-modal classification networks hard?
\newblock In \emph{Proceedings of Conference on Computer Vision and Pattern
  Recognition (CVPR)}, 2020{\natexlab{a}}.

\bibitem[Wang et~al.(2020{\natexlab{b}})Wang, Tran, and Feiszli]{wang2020makes}
Weiyao Wang, Du~Tran, and Matt Feiszli.
\newblock What makes training multi-modal classification networks hard?
\newblock In \emph{Proceedings of Conference on Computer Vision and Pattern
  Recognition (CVPR)}, 2020{\natexlab{b}}.

\bibitem[Wang et~al.(2016{\natexlab{b}})Wang, Farhadi, and
  Gupta]{Wang_Transformation}
Xiaolong Wang, Ali Farhadi, and Abhinav Gupta.
\newblock Actions {\textasciitilde} transformations.
\newblock In \emph{Proceedings of Conference on Computer Vision and Pattern
  Recognition (CVPR)}, 2016{\natexlab{b}}.

\bibitem[Wray et~al.(2019)Wray, Larlus, Csurka, and Damen]{Wray_2019_ICCV}
Michael Wray, Diane Larlus, Gabriela Csurka, and Dima Damen.
\newblock Fine-grained action retrieval through multiple parts-of-speech
  embeddings.
\newblock In \emph{Proceedings of International Conference on Computer Vision
  (ICCV)}, 2019.

\bibitem[Wu et~al.(2019)Wu, Feichtenhofer, Fan, He, Krahenbuhl, and
  Girshick]{Wu_2019_CVPR}
Chao-Yuan Wu, Christoph Feichtenhofer, Haoqi Fan, Kaiming He, Philipp
  Krahenbuhl, and Ross Girshick.
\newblock Long-term feature banks for detailed video understanding.
\newblock In \emph{Proceedings of Conference on Computer Vision and Pattern
  Recognition (CVPR)}, 2019.

\bibitem[Xiao et~al.(2020{\natexlab{a}})Xiao, Lee, Grauman, Malik, and
  Feichtenhofer]{xiao2020audiovisual}
Fanyi Xiao, Yong~Jae Lee, Kristen Grauman, Jitendra Malik, and Christoph
  Feichtenhofer.
\newblock Audiovisual slowfast networks for video recognition.
\newblock \emph{arXiv preprint arXiv:2001.08740}, 2020{\natexlab{a}}.

\bibitem[Xiao et~al.(2020{\natexlab{b}})Xiao, Lee, Grauman, Malik, and
  Feichtenhofer]{xiao_2020_arxiv}
Fanyi Xiao, Yong~Jae Lee, Kristen Grauman, Jitendra Malik, and Christoph
  Feichtenhofer.
\newblock Audiovisual slowfast networks for video recognition.
\newblock \emph{CoRR}, abs/2001.08740, 2020{\natexlab{b}}.

\bibitem[Yang et~al.(2019)Yang, Dai, Yang, Carbonell, Salakhutdinov, and
  Le]{yang2019xlnet}
Zhilin Yang, Zihang Dai, Yiming Yang, Jaime Carbonell, Russ~R Salakhutdinov,
  and Quoc~V Le.
\newblock {XLN}et: Generalized autoregressive pretraining for language
  understanding.
\newblock In \emph{Advances in Neural Information Processing Systems
  (NeurIPS)}, 2019.

\bibitem[Ye et~al.(2019)Ye, Rochan, Liu, and Wang]{ye2019cross}
Linwei Ye, Mrigank Rochan, Zhi Liu, and Yang Wang.
\newblock Cross-modal self-attention network for referring image segmentation.
\newblock In \emph{Proceedings of Conference on Computer Vision and Pattern
  Recognition (CVPR)}, 2019.

\bibitem[Zhang et~al.(2021)Zhang, Gupta, and Zisserman]{zhang_2021_CVPR}
Chuhan Zhang, Ankush Gupta, and Andrew Zisserman.
\newblock Temporal query networks for fine-grained video understanding.
\newblock In \emph{Proceedings of Conference on Computer Vision and Pattern
  Recognition (CVPR)}, 2021.

\bibitem[Zhang et~al.(2018)Zhang, Ciss{\'{e}}, Dauphin, and
  Lopez{-}Paz]{Zhang_2018_ICLR}
Hongyi Zhang, Moustapha Ciss{\'{e}}, Yann~N. Dauphin, and David Lopez{-}Paz.
\newblock mixup: Beyond empirical risk minimization.
\newblock In \emph{Proceedings of International Conference on Learning
  Representations {ICLR}}, 2018.

\bibitem[Zhou et~al.(2018)Zhou, Andonian, Oliva, and
  Torralba]{zhou2018temporal}
Bolei Zhou, Alex Andonian, Aude Oliva, and Antonio Torralba.
\newblock Temporal relational reasoning in videos.
\newblock In \emph{Proceedings of European Conference on Computer Vision
  (ECCV)}, 2018.

\end{thebibliography}

\clearpage
\section*{Appendix}
\appendix
\section{EPIC-KITCHENS-100: Results on the Test Set}
\label{sec:epic-kitchens-100}

\begin{table}[t]
    \centering
\resizebox{\textwidth}{!}{%
\begin{tabular}{@{}lrrrrrrrrrrrr@{}}
\toprule
                                                                                     & \multicolumn{6}{c}{Overall}                           & \multicolumn{3}{c}{Unseen Participants} & \multicolumn{3}{c}{Tail-classes} \\
                                                                                      \cmidrule(r){2-7}                                       \cmidrule(lr){8-10}                   \cmidrule(l){11-13}
                                                                                  &   \multicolumn{3}{c}{Top-1 Accuracy (\%)} & \multicolumn{3}{c}{Top-5 Accuracy (\%)} & \multicolumn{3}{c}{Top-1 Accuracy (\%)}           & \multicolumn{3}{c}{Top-1 Accuracy (\%)}    \\
                                                                                      \cmidrule(r){2-4}           \cmidrule(lr){5-7}          \cmidrule(lr){8-10}                   \cmidrule(l){11-13}
                                        Model                                    & \multicolumn{1}{c}{Verb} & \multicolumn{1}{c}{Noun} & \multicolumn{1}{c}{Action} & \multicolumn{1}{c}{Verb} & \multicolumn{1}{c}{Noun} & \multicolumn{1}{c}{Action}& \multicolumn{1}{c}{Verb} & \multicolumn{1}{c}{Noun} & \multicolumn{1}{c}{Action} & \multicolumn{1}{c}{Verb} & \multicolumn{1}{c}{Noun} & \multicolumn{1}{c}{Action} \\ \midrule
                                           
										  TSN~\cite{TSN2016ECCV}                  & 59.03 & 46.78 & 33.57 & 87.55 & 72.10 & 53.89 & 53.11 & 42.02 & 27.37 & 26.23 & 14.73 & 11.43 \\
                                          TBN~\cite{kazakos2019TBN}                & 62.72 & 47.59 & 35.48 & 88.77 & 73.08 & 56.34 & 56.69 & 43.65 & 29.27 & 30.97 & 19.52 & 14.10 \\
                                          TSM~\cite{lin2019tsm}                & 65.32 & 47.80 & 37.39 & 89.16 & 73.95 & 57.89 & 59.68 & 42.51 & 30.61 & 30.03 & 16.96 & 13.45 \\
                                          SlowFast~\cite{Feichtenhofer_2019_ICCV}  & 63.79 & 48.55 & 36.81 & 88.84 & 74.49 & 56.39 & 57.66 & 42.55 & 29.27 & 29.65 & 17.11 & 13.45 \\
										  Ego-Exo~\cite{Li_2021_CVPR} 	        & 66.07 & 51.51 & 39.98 & \textbf{89.39} & 76.31 & 60.68 & 59.83 & 45.50 & 32.63 & 33.92 & 22.91 & 16.96\\
                                          MTCN - v.f. SlowFast~\cite{Feichtenhofer_2019_ICCV}                                     & \textbf{68.44} & 55.41 & 44.10 & 88.74 & 78.04 & 61.69 & \textbf{61.82} & 47.62 & 34.94 & 34.77 & 28.60 & 20.45\\
                                        MTCN - v.f. Mformer-HR~\cite{patrick_2021_arXiv} & 67.88 & \textbf{60.02} & \textbf{46.83} & 88.69 & \textbf{81.84} & \textbf{66.48} & 61.07 & \textbf{55.21} & \textbf{38.98} & \textbf{35.16} & \textbf{34.70} & \textbf{22.79}\\ \bottomrule
\end{tabular}}
    \caption{Results on the test set of EPIC-Kitchens-100.}
    \label{tab:test_set}
\end{table}

\begin{table}[t]
    \centering
\resizebox{0.7\textwidth}{!}{%
\begin{tabular}{@{}lrrrrrr@{}}
\toprule
                                        & \multicolumn{3}{c}{Top-1 Accuracy (\%)} & \multicolumn{3}{c}{Top-5 Accuracy (\%)}\\
                                                                                      \cmidrule(r){2-4}           \cmidrule(lr){5-7}
                                      Model                                    & \multicolumn{1}{c}{Verb} & \multicolumn{1}{c}{Noun} & \multicolumn{1}{c}{Action} & \multicolumn{1}{c}{Verb} & \multicolumn{1}{c}{Noun} & \multicolumn{1}{c}{Action}\\ \midrule
                                            LFB~\cite{Wu_2019_CVPR}            & 60.0 & 45.0 & 32.7 & 88.4 & 71.8 & 55.3\\
                                            G-Blend~\cite{wang2020makes} 	    & 66.7 & 48.5 & 37.1 & 88.9 & 71.4 & 56.2\\
            								 AV-SlowFast~\cite{xiao2020audiovisual}\hspace{36pt} & 65.7 & 46.4 & 35.9 & 89.5 & 71.7 & 57.8\\
            								 Ego-Exo~\cite{Li_2021_CVPR}    & 65.97 & 47.99 & 37.09 & \textbf{90.32} & 70.72 & 56.32\\
            							MTCN (Ours)                               & \textbf{69.12} & \textbf{51.30} & \textbf{40.77} & 90.18 & \textbf{73.53} & \textbf{59.15} \\
            								
            								\bottomrule
\end{tabular}}
    \caption{Comparison with SOTA on the Seen split (S1) of EPIC-KITCHENS-55.}
    \label{tab:epic_55}

\end{table}

In Table~\ref{tab:sota} of the main paper, we compare to published works on the validation set of EPIC-KITCHENS-100.
Unfortunately, most works do not report on the leaderboard test set.
In Table~\ref{tab:test_set}, we provide results on the test set comparing our model to baselines from \cite{Damen2020RESCALING}, as well as Ego-Exo~\cite{Li_2021_CVPR} that distills knowledge from a much larger training set. MTCN outperforms all other methods, including the competitive method of~\cite{Li_2021_CVPR}, showcasing that multimodal temporal context from consecutive actions is more beneficial than pretraining large models (ResNet101) using egocentric signals from  third-person datasets. 

\section{EPIC-KITCHENS-55 Results}
\label{sec:epic-kitchens-55}

We also compare our model to works that report on the earlier version of this dataset, namely EPIC-KITCHENS-55~\cite{Damen2018EPICKITCHENS} in Table~\ref{tab:epic_55}. We report results for the Seen split (S1). We opted to include these in the appendix to avoid confusion in the main paper as the results are not comparable across these two dataset versions. We compare MTCN with two audio-visual approaches \cite{wang2020makes} and \cite{xiao2020audiovisual},  as well as \cite{Wu_2019_CVPR} which  was one of the first works to utilise temporal context. 
We also report the performance of \cite{Li_2021_CVPR} which evaluates their method on both EPIC-KITCHENS-55 and EPIC-KITCHENS-100.
Our MTCN outperforms all approaches.

\section{Language model analysis and baselines}
\label{sec:lm_analysis}

In this section, we assess the statistical significance of our language model and compare the performance of our MTCN to variants using baseline language models. All the experiments in this section as well as in Sections~\ref{sec:online} and~\ref{sec:ablation} and the visualisations in Section~\ref{sec:attention} are performed on the validation set of EPIC-KITCHENS-100 and using the SlowFast visual features.

\noindent \noindent \textbf{Statistical significance of LM}. We train 10 audio-visual transformers and 10 corresponding language models with different random seeds. Table~\ref{tab:multiplerun} shows the mean and standard deviation top-1 and top-5 accuracy without and with the language model. Utilising the language model improves performance on average with a low std, demonstrating that the improvement from the language model is statistically significant. We further showcase that by conducting T-tests on verb, noun and action top-1 accuracies, obtaining a p-value of $3.6e-2$, $6.0e-4$, $9.7e-4$, respectively.

\begin{table}[t]
    \centering
\resizebox{0.8\textwidth}{!}{%
\begin{tabular}{@{}>{\color{black}}c>{\color{black}}r>{\color{black}}r>{\color{black}}r>{\color{black}}r>{\color{black}}r>{\color{black}}r@{}}
\toprule
                                        & \multicolumn{3}{c}{Top-1 Accuracy (\%)} & \multicolumn{3}{c}{Top-5 Accuracy (\%)}\\
                                                                                      \cmidrule(r){2-4}           \cmidrule(lr){5-7}
                                      LM                                    & \multicolumn{1}{c}{Verb} & \multicolumn{1}{c}{Noun} & \multicolumn{1}{c}{Action} & \multicolumn{1}{c}{Verb} & \multicolumn{1}{c}{Noun} & \multicolumn{1}{c}{Action}\\ \midrule
                                            	\xmark	            & 70.26 $\pm$ 0.27 & 55.70 $\pm$ 0.22 & 44.90 $\pm$ 0.20 & 91.12 $\pm$ 0.13 & \textbf{79.03} $\pm$ \textbf{0.18} & \textbf{64.79} $\pm$ \textbf{0.17}\\
                                                \cmark          & \textbf{70.52} $\pm$ \textbf{0.25} & \textbf{56.08} $\pm$ \textbf{0.21} & \textbf{45.25} $\pm$ \textbf{0.18} & \textbf{91.13} $\pm$ \textbf{0.13} & \textbf{79.03} $\pm$ \textbf{0.18} & 64.58 $\pm$ 0.18\\
            								\bottomrule
\end{tabular}}
    \caption{Mean and standard deviation of multiple runs both w. \& w/o language model in the validation set of EPIC-KITCHENS-100.}
    \label{tab:multiplerun}
\end{table}

\begin{table}[t]
    \centering
\resizebox{\textwidth}{!}{%
\begin{tabular}{@{}>{\color{black}}l>{\color{black}}r>{\color{black}}r>{\color{black}}r>{\color{black}}r>{\color{black}}r>{\color{black}}r>{\color{black}}r>{\color{black}}r>{\color{black}}r>{\color{black}}r>{\color{black}}r>{\color{black}}r@{}}
\toprule
                                                                 & \multicolumn{6}{c}{Overall}                           & \multicolumn{3}{c}{Unseen Participants} & \multicolumn{3}{c}{Tail-classes} \\
                                                                  \cmidrule(r){2-7}                                       \cmidrule(lr){8-10}                   \cmidrule(l){11-13}
                                                                 & \multicolumn{3}{c}{Top-1 Accuracy (\%)} & \multicolumn{3}{c}{Top-5 Accuracy (\%)} & \multicolumn{3}{c}{Top-1 Accuracy (\%)}           & \multicolumn{3}{c}{Top-1 Accuracy (\%)}    \\
                                                                  \cmidrule(r){2-4}           \cmidrule(lr){5-7}          \cmidrule(lr){8-10}                   \cmidrule(l){11-13}
                  Model                                   & \multicolumn{1}{c}{Verb} & \multicolumn{1}{c}{Noun} & \multicolumn{1}{c}{Action} & \multicolumn{1}{c}{Verb} & \multicolumn{1}{c}{Noun} & \multicolumn{1}{c}{Action}& \multicolumn{1}{c}{Verb} & \multicolumn{1}{c}{Noun} & \multicolumn{1}{c}{Action} & \multicolumn{1}{c}{Verb} & \multicolumn{1}{c}{Noun} & \multicolumn{1}{c}{Action} \\ \midrule

                        No LM & 70.23 & 55.82 & 45.00 & 91.13 & \textbf{79.06} & \textbf{64.58} & 63.29 & 46.38 & 35.02 & \textbf{41.76} & 32.26 & 24.41\\
                        \cmidrule{1-13}\morecmidrules\cmidrule{1-13}
						N-gram	& 70.23 & 55.84 & 45.02 & 91.13 & \textbf{79.06} & 64.49 & 63.29 & 46.38 & 35.02 & \textbf{41.76} & 32.26 & 24.41\\
						BiLSTM & 70.57 & 55.97 & 45.09 & \textbf{91.14} & \textbf{79.06} & \textbf{64.55} & 63.29 & 46.76 & 35.31 & 40.68 & 32.47 & 24.15\\
						Transformer enc.~(proposed)   & \textbf{70.60} & \textbf{56.26} & \textbf{45.48} & \textbf{91.14} & \textbf{79.06} & 63.06 & \textbf{63.76} & \textbf{47.14} & \textbf{35.87} & 41.36 & \textbf{32.84} & \textbf{24.70}\\\bottomrule
\end{tabular}}
    \caption{Performance of MTCN in the validation set of EPIC-KITCHENS-100 using different language models.}
    \label{tab:architecture_LM}
\end{table}

\noindent \noindent \textbf{Baselines comparison}. We compare our MTCN that uses a transformer based language model to two baselines, N-gram and Bi-directional LSTM (BiLSTM). For N-gram, we follow a similar procedure to natural language processing. In particular, from all action sequences of length 9 in the training set, we derive the heuristic probability of occurrence of the centre action given the preceding and succeeding actions. We train a BiLSTM with 3 layers and a hidden size of 512. The rest hyperparameters are the same as the transformer encoder.

Results are shown in Table~\ref{tab:architecture_LM}. It turns out that only a few preceding-succeeding action sequences in the training set also appear in the validation set, resulting in no difference in performance when N-gram is added comparing to not using a language model. Our transformer-based Masked Language Model (MLM) outperforms both the N-gram and BiLSTM, showcasing that it is beneficial to use a deep neural network language model over a heuristic prior and that MLM with transformers outperforms recurrent architectures in this problem.


\section{Online recognition}
\label{sec:online}

The focus of this work is to leverage both past and future context to predict an action. In this section however, we explore the performance of our model in online recognition, i.e.\ using only the preceding actions as context to predict the current action. This approach can be used to recognise actions in an online fashion for streaming videos. For this setting, we train the audio-visual transformer to predict the last action in the sequence. We do not train a new language model for this task; we simply mask and predict the last action in the sequence instead of the centre one.
 
Results are demonstrated in Table~\ref{tab:online_recognition} by varying $w$. 
Our model can also utilise temporal context in this setting, as performance improves for $w>1$ with optimal top-1 accuracy at $w=7$ and optimal accuracy on tail-classes at $w=9$. Compared to our original proposal that utilises also future context (see Table~\ref{tab:temporal_context_extent} on main paper), the overall performance degrades, indicating that leveraging future context is beneficial.

\begin{table*}[t]
    \centering
\resizebox{\textwidth}{!}{%
\begin{tabular}{@{}>{\color{black}}l>{\color{black}}r>{\color{black}}r>{\color{black}}r>{\color{black}}r>{\color{black}}r>{\color{black}}r>{\color{black}}r>{\color{black}}r>{\color{black}}r>{\color{black}}r>{\color{black}}r>{\color{black}}r@{}}
\toprule
                                                                                     & \multicolumn{6}{c}{Overall}                           & \multicolumn{3}{c}{Unseen Participants} & \multicolumn{3}{c}{Tail-classes} \\
                                                                                      \cmidrule(r){2-7}                                       \cmidrule(lr){8-10}                   \cmidrule(l){11-13}
                                                                                     & \multicolumn{3}{c}{Top-1 Accuracy (\%)} & \multicolumn{3}{c}{Top-5 Accuracy (\%)} & \multicolumn{3}{c}{Top-1 Accuracy (\%)}           & \multicolumn{3}{c}{Top-1 Accuracy (\%)}    \\
                                                                                      \cmidrule(r){2-4}           \cmidrule(lr){5-7}          \cmidrule(lr){8-10}                   \cmidrule(l){11-13}
                                      $w$                                    & \multicolumn{1}{c}{Verb} & \multicolumn{1}{c}{Noun} & \multicolumn{1}{c}{Action} & \multicolumn{1}{c}{Verb} & \multicolumn{1}{c}{Noun} & \multicolumn{1}{c}{Action}& \multicolumn{1}{c}{Verb} & \multicolumn{1}{c}{Noun} & \multicolumn{1}{c}{Action} & \multicolumn{1}{c}{Verb} & \multicolumn{1}{c}{Noun} & \multicolumn{1}{c}{Action} \\ \midrule
                                            1	& 67.93 & 52.29 & 41.30 & 90.53 & 76.47 & 61.52 & 61.13 & 44.60 & 32.58 & 42.05 & 27.42 & 21.48\\
            								3 & 68.42 & 54.15 & 42.59 & 91.20 & 78.52 & 61.10 & \textbf{61.69} & \textbf{44.41} & \textbf{32.11} & 40.11 & 31.26 & 22.58\\
            								5 & 68.58 & 54.04 & 42.75 & \textbf{90.96} & 78.27 & 62.04 & 59.81 & 43.94 & \textbf{32.11} & 39.49 & 31.11 & 22.48\\
            								7 & \textbf{68.88} & \textbf{54.31} & \textbf{42.96} & 90.89 & 77.87 & 62.39 & 61.41 & 43.38 & 32.02 & 40.51 & 32.00 & 23.61\\
            								9 &  68.77 & 54.28 & 42.77 & 90.66 & 77.72 & \textbf{62.44} & 60.38 & 45.07 & 31.83 & \textbf{40.80} & \textbf{32.68} & \textbf{23.86}\\
            								11 & 67.83 & 54.04 & 42.13 & 90.63 & \textbf{78.85} & 62.10 & 57.46 & 43.94 & 31.46 & 36.88 & 30.74 & 21.96\\
            								\bottomrule
            								
\end{tabular}}
    \caption{Online action recognition results by varying temporal context length in the validation set of EPIC-KITCHENS-100.}
    \label{tab:online_recognition}
\end{table*}

\begin{table}[t!]
    \centering
\resizebox{\textwidth}{!}{%
\begin{tabular}{@{}>{\color{black}}l>{\color{black}}c>{\color{black}}r>{\color{black}}r>{\color{black}}r>{\color{black}}r>{\color{black}}r>{\color{black}}r>{\color{black}}r>{\color{black}}r>{\color{black}}r>{\color{black}}r>{\color{black}}r>{\color{black}}r@{}}
\toprule
                                                                 && \multicolumn{6}{c}{Overall}                           & \multicolumn{3}{c}{Unseen Participants} & \multicolumn{3}{c}{Tail-classes} \\
                                                                  \cmidrule(r){3-8}                                       \cmidrule(lr){9-11}                   \cmidrule(l){12-14}
                                                                 && \multicolumn{3}{c}{Top-1 Accuracy (\%)} & \multicolumn{3}{c}{Top-5 Accuracy (\%)} & \multicolumn{3}{c}{Top-1 Accuracy (\%)}           & \multicolumn{3}{c}{Top-1 Accuracy (\%)}    \\
                                                                  \cmidrule(r){3-5}           \cmidrule(lr){6-8}          \cmidrule(lr){9-11}                   \cmidrule(l){12-14}
                  Layers    &   Shared                              & \multicolumn{1}{c}{Verb} & \multicolumn{1}{c}{Noun} & \multicolumn{1}{c}{Action} & \multicolumn{1}{c}{Verb} & \multicolumn{1}{c}{Noun} & \multicolumn{1}{c}{Action}& \multicolumn{1}{c}{Verb} & \multicolumn{1}{c}{Noun} & \multicolumn{1}{c}{Action} & \multicolumn{1}{c}{Verb} & \multicolumn{1}{c}{Noun} & \multicolumn{1}{c}{Action} \\ \midrule

                                    1 & - & 69.58 & 55.04 & 43.71 & 91.27 & 79.02 & 63.96 & 61.03 & 46.01 & 33.33 & 42.10 & 32.42 & 24.09 \\
                                    \cmidrule(r){1-14}
                                    2	&	\xmark	& 69.49 & 55.41 & 43.94 & \textbf{91.13} & 78.96 & 63.86 & \textbf{62.72} & 46.57 & 34.37 & \textbf{41.42} & \textbf{32.32} & 23.90 \\
                                    4	&	\xmark	& \textbf{71.01} & \textbf{56.55} & \textbf{46.04} & 90.98 & \textbf{79.28} & \textbf{63.97} & 62.35 & \textbf{47.61} & \textbf{35.96} & 39.94 & 31.89 & \textbf{24.22} \\
                                    6	&	\xmark	& 69.58 & 55.68 & 44.89 & 90.28 & 78.17 & 63.37 & 61.31 & 45.63 & 34.46 & 38.75 & 32.21 & 23.90 \\
                                    \cmidrule(r){1-14}
                                    2	&	\cmark  & 69.82 & 55.37 & 43.81 & 91.09 & 78.97 & \textbf{64.26} & 61.50 & 44.32 & 32.68 & \textbf{42.05} & 32.58 & 23.74 \\
                                    4	&	\cmark	& \textbf{70.60} & \textbf{56.26} & \textbf{45.48} & \textbf{91.14} & \textbf{79.06} & 63.06 & \textbf{63.76} & \textbf{47.14} & \textbf{35.87} & 41.36 & \textbf{32.84} & \textbf{24.70} \\
                                    6	&	\cmark	& 69.58 & 55.68 & 44.89 & 90.28 & 78.17 & 63.37 & 61.31 & 45.63 & 34.46 & 38.75 & 32.21 & 23.90 \\\bottomrule
\end{tabular}}
    \caption{Analysis of performance using different number of layers, both w. and w/o weight sharing. Results are shown in the validation set of EPIC-KITCHENS-100.}
    \label{tab:num_layers}
\end{table}

\section{Architecture ablations}
\label{sec:ablation}

\begin{table}[t]
    \centering
\resizebox{\textwidth}{!}{%
\begin{tabular}{@{}lrrrrrrrrrrrr@{}}
\toprule
                                                                                     & \multicolumn{6}{c}{Overall}                           & \multicolumn{3}{c}{Unseen Participants} & \multicolumn{3}{c}{Tail-classes} \\
                                                                                      \cmidrule(r){2-7}                                       \cmidrule(lr){8-10}                   \cmidrule(l){11-13}
                                                                                     & \multicolumn{3}{c}{Top-1 Accuracy (\%)} & \multicolumn{3}{c}{Top-5 Accuracy (\%)} & \multicolumn{3}{c}{Top-1 Accuracy (\%)}           & \multicolumn{3}{c}{Top-1 Accuracy (\%)}    \\
                                                                                      \cmidrule(r){2-4}           \cmidrule(lr){5-7}          \cmidrule(lr){8-10}                   \cmidrule(l){11-13}
                                      Pos. enc.                                    & \multicolumn{1}{c}{Verb} & \multicolumn{1}{c}{Noun} & \multicolumn{1}{c}{Action} & \multicolumn{1}{c}{Verb} & \multicolumn{1}{c}{Noun} & \multicolumn{1}{c}{Action}& \multicolumn{1}{c}{Verb} & \multicolumn{1}{c}{Noun} & \multicolumn{1}{c}{Action} & \multicolumn{1}{c}{Verb} & \multicolumn{1}{c}{Noun} & \multicolumn{1}{c}{Action} \\ \midrule

            								Fourier PE	            & 69.60 & 56.13 & 44.63 & 90.65 & 78.86 & 63.31 & 63.29 & 45.63 & 35.12 & 38.86 & 32.53 & 23.09\\
            								Relative PE             &70.32 & \textbf{56.30} & 45.37 & 91.01 & \textbf{79.35} & \textbf{64.04} & 61.41 & 46.67 & 33.99 & \textbf{41.42} & \textbf{33.74} & \textbf{24.96} \\
            								Absolute PE (Proposed)  & \textbf{70.60} & 56.26 & \textbf{45.48} & \textbf{91.14} & 79.06 & 63.06 & \textbf{63.76} & \textbf{47.14} & \textbf{35.87} & 41.36 & 32.84 & 24.70\\\bottomrule
\end{tabular}}
    \caption{Comparison of different positional encodings (PE) using the validation set of EPIC-KITCHENS-100.}
    \label{tab:pos_enc}
\end{table}

In Table~\ref{tab:num_layers}, we explore different number of layers in MTCN, both without and with (layer-wise) weight sharing, and compare each case with a single layer. Note that we use the same number of layers and sharing strategy for both AV and LM.
We use bold to indicate best performance within each group rather than overall. Best results are obtained using four layers in most metrics, both without \& with weight sharing. 
These outperform a single layer, demonstrating that is beneficial to use a multi-layered transformer. 
Although MTCN without weight sharing performs slightly better, our proposed model has $2.7\times$ less parameters with only a minor drop in performance. 

In Table~\ref{tab:pos_enc}, we compare the effect of different types of positional encodings. Particularly, we replace our chosen absolute learnt positional encoding with relative positional encodings \cite{shaw_2018_naacl} and Fourier feature positional encodings \cite{jaegle2021perceiver}. Fourier feature positional encodings replace our learnable absolute positional encodings with non-learnable ones represented as a vector of log-linearly spaced frequency bands up to a maximum frequency. Relative positional encodings replace our absolute positional encodings of the inputs, with positional encodings representing distances between tokens and placed within the self-attention layers. As shown in the table our proposed absolute learnable positional encodings outperform Fourier feature positional encodings in all metrics (except top-5 action accuracy). Comparing to relative positional encodings, our positional encodings are slightly better in top-1 verbs and actions, as well as in unseen participants, while relative positional encodings perform slightly better in top-5 accuracy and tail classes. Overall, there are no notable differences between the different choices of positional encodings.


\section{EGTEA Implementation Details}
\label{sec:egtea_details}

\noindent\textbf{Visual features}. For EGTEA, we train SlowFast~\cite{Feichtenhofer_2019_ICCV} using the EPIC-KITCHENS pre-trained model, by sampling a clip of 2s from an action segment similar to EPIC-KITCHENS. We use a learning rate of 0.001, no warm-up, and we keep the batch normalisation layers frozen. All unspecified hyperparameters remain unchanged. For feature extraction, we follow the same procedure as EPIC-KITCHENS, except that we use clips of 2s rather than 1s.

\noindent \noindent\textbf{Train/Val Details}. Here, we discuss differences in the architecture for training/evaluating EGTEA. Remember that for EGTEA we train only vision and language as EGTEA does not contain audio. First, as there is no audio input to the transformer, we do not use modality encodings either. Second, following previous methods 
\cite{LiYin_2018_ECCV,Min_2021_WACV,Lu_2019_ICCV,Sudhakaran_2018_BMVC,Sudhakaran_2019_CVPR,Huang_2020_TIP} that train using a single head for actions and report only action accuracy, we use a single summary embedding for actions, rather than verb/noun embeddings. Accordingly, the language model utilises a single word-embedding for actions, with a dimension of 512. For training the visual-only transformer, we use a learning rate of 0.001, train the model for 50 epochs and decay the learning rate at epochs 25 and 38, while keeping all other hyperparameters unchanged. We use same hyperparameters for the language model. For evaluation, differently than EPIC-KITCHENS, we average the predictions of the 10 clips per action, rather than feeding all 10 clips in the transformer.

\section{Ablation of temporal context and language model in EGTEA}
\label{sec:egtea_ablation}

\begin{table}[t]
    \centering
\resizebox{0.7\textwidth}{!}{%
\begin{tabular}{@{}>{\color{black}}l>{\color{black}}r>{\color{black}}r>{\color{black}}r>{\color{black}}r@{}}
\toprule
          & \multicolumn{2}{c}{Visual} & \multicolumn{2}{c}{Visual + LM} \\
          \cmidrule(r){2-3} \cmidrule(l){4-5}                  
        $w$ & Top-1(\%)    & Mean Class (\%)   & Top-1(\%)    & Mean Class (\%)     \\
        \midrule
        1 & 72.26   & 64.98            & 72.26       & 64.98     \\
        3 & 72.55   & 64.86            & \textbf{73.59}      & 65.87              \\
        5 & \textbf{73.10}   & \textbf{65.42}            & 73.49      & 65.57              \\
        7 & 72.26   & 64.38            & 73.19      & 65.31              \\
        9 & 72.55   & 64.86            & 73.44      & \textbf{66.02}\\\bottomrule             
\end{tabular}}
\caption{Ablation of temporal context extent and language model in the first test split of EGTEA.}
\label{tab:egtea_ablation}
\end{table}

\begin{figure}[t!]
    \centering
    \includegraphics[width=\textwidth]{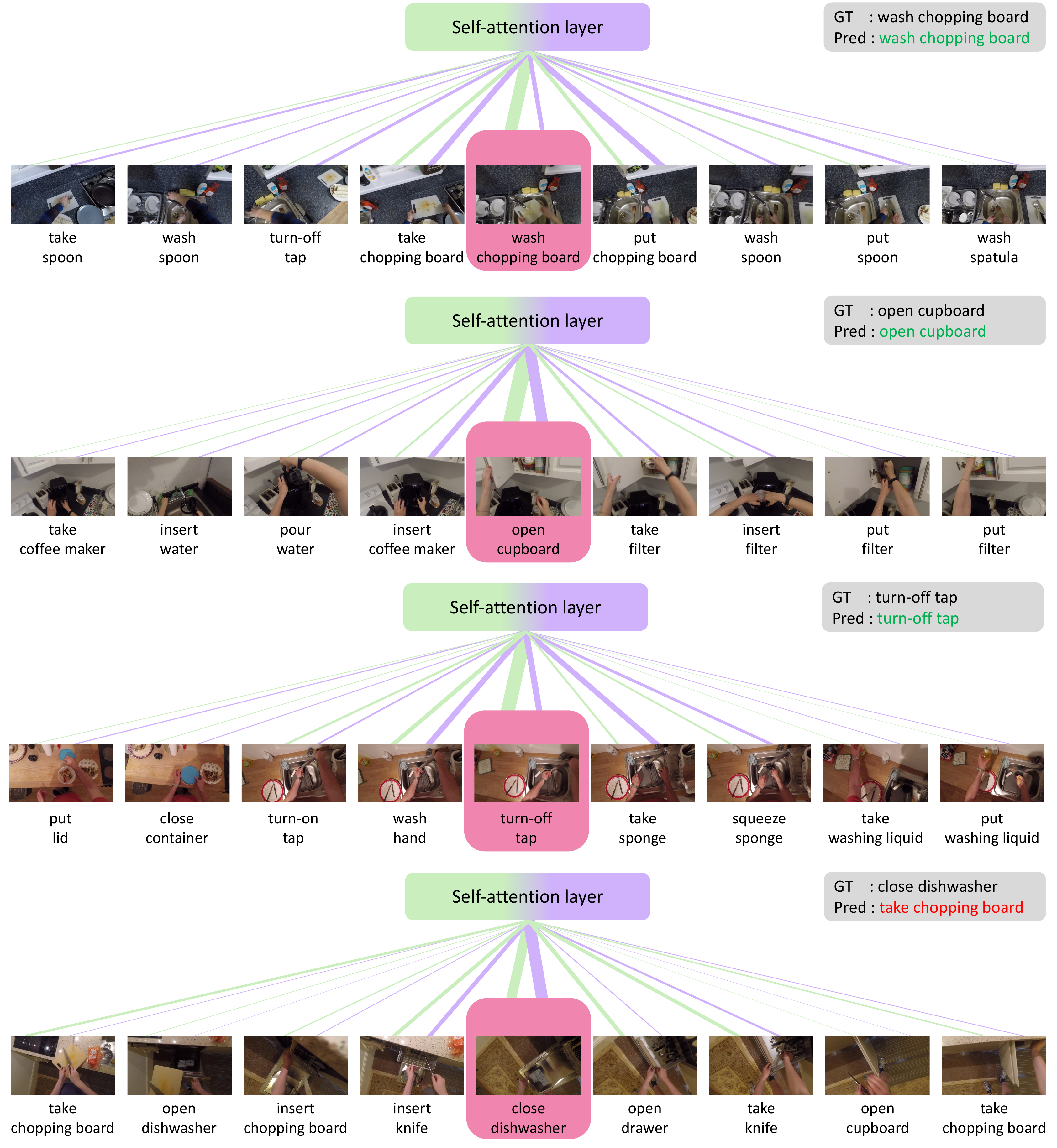}
    \caption{Additional qualitative results of attention weights along with the predictions of our model. Green and purple edges represent attention to visual and auditory tokens, respectively, from the noun summary embedding. Thickness indicates attention weights magnitude to centre (bordered) and temporal context actions.} 
    \label{fig:more_attention_examples}
    \vspace*{-8pt}
\end{figure}

We study the effect of the temporal context length both with and without the language model on the first test split of EGTEA. Results are shown in Table~\ref{tab:egtea_ablation}. For the visual only model, top-1 accuracy increases when we increase the length of temporal context from $w=1$ to 5, and optimal results for both top-1 and mean class accuracy are obtained for $w=5$. When the language model is incorporated top-1 accuracy increases from $w=1$ to 3 and then decreases while best mean class accuracy is obtained at $w=9$. These findings showcase that our model successfully utilises context in this dataset as well. The language model is helpful for EGTEA, and provides a bigger boost in performance than EPIC-KITCHENS, possibly due to the absence of audio modality. Finally, it is worth noting that after the addition of the language model best performance is obtained at a shorter temporal context, showing that shorter sequences of actions provide a stronger prior in this dataset.

\section{Attention Visualisation}
\label{sec:attention}

In Fig.~\ref{fig:more_attention_examples}, we show additional qualitative examples, similar to Fig.~\ref{fig:att_visualisation} in the main paper. These demonstrate how our model attends to temporal context. 
In the first three examples, the model predicts the centre action correctly, while in the last one it gives incorrect predictions. In the `wash chopping board' example, the model particularly attends to actions containing the chopping board. For `open cupboard', the model has high audio-visual attention to the centre action, and high attention to the audio of the previous action (`insert coffee maker'), showing that at times audio provides useful temporal context. The importance of audio is also apparent in the third example. A source of error in the model results from confusing the centre action with another action in the sequence; in the fourth example `close dishwasher' is predicted as `take chopping board' which corresponds to the first and last actions in the temporal context.

\end{document}